\pdfoutput=1

\documentclass[11pt]{article}

\usepackage[preprint]{acl}

\usepackage{times}
\usepackage{latexsym}

\usepackage[T1]{fontenc}

\usepackage[utf8]{inputenc}

\usepackage{microtype}

\usepackage{inconsolata}

\usepackage{graphicx}

\usepackage{indentfirst}
\usepackage{xspace}
\usepackage{tabularx}
\usepackage{multirow}
\usepackage[capitalize,noabbrev,nameinlink]{cleveref}
\usepackage{xcolor}  
\usepackage{array} 
\usepackage{booktabs}
\usepackage{colortbl}
\usepackage{enumitem}
\usepackage{threeparttable}
\usepackage{tablefootnote}

\definecolor{darkgreen}{rgb}{0.0, 0.5, 0.0} 
\newcommand{\upgreen}{\textcolor{darkgreen}{$\uparrow$}}

\newcommand{\downred}{\textcolor{red}{$\downarrow$}} 
\newcommand{\TPO}{{ThinkPO}\xspace}

\title{Thinking Preference Optimization}

\author{Wang Yang, Hongye Jin, Jingfeng Yang, Vipin Chaudhary, Xiaotian Han \\
\{wxy320, vxc204, xhan\}@case.edu; jhy0410@tamu.edu; jingfengyangpku@gmail.com}

\begin{document}

\maketitle

\begin{abstract}

Supervised Fine-Tuning (SFT) has been a go-to and effective method for enhancing long chain-of-thought (CoT) reasoning in relatively small LLMs by fine-tuning them with long CoT responses from larger LLMs~\footnote{Deepseek official distilled models  \href{https://huggingface.co/collections/deepseek-ai/deepseek-r1-678e1e131c0169c0bc89728d}{DeepSeek-R1-Distill}, \href{https://huggingface.co/open-thoughts/OpenThinker-7B}{OpenThinker-7B}, \href{https://huggingface.co/NovaSky-AI/Sky-T1-32B-Preview}{Sky-T1-32B}, and \href{https://huggingface.co/bespokelabs/Bespoke-Stratos-7B}{Bespoke-Stratos-7B} was trained in this way. }. To continually improve reasoning abilities, we can either collect new high-quality long CoT reasoning SFT data or repeatedly train on existing SFT datasets. However, acquiring new long CoT SFT data is costly and limited, while repeated training often results in a performance plateau or decline. To further boost the performance with the SFT data, we propose Thinking Preference Optimization (\TPO), a simple yet effective post-SFT method that enhances long CoT reasoning without requiring new long CoT responses.  Instead, \TPO~utilizes readily available or easily obtainable short CoT reasoning responses as rejected answers and long CoT responses as chosen answers for the same question. It then applies direct preference optimization to encourage the model to favor longer reasoning outputs. Experiments show that \TPO~further improves the reasoning performance of SFT-ed models, e.g. it increases math reasoning accuracy of SFT-ed models by $8.6\%$ and output length by $25.9\%$. Notably, \TPO~is capable of continually boosting the performance of the publicly distilled SFT model, e.g., increasing the official DeepSeek-R1-Distill-Qwen-7B's performance on MATH500 from $87.4\%$ to $91.2\%$. Our code is available at \url{https://github.com/uservan/ThinkPO}.
 
\end{abstract}

\begin{figure}[t]
\includegraphics[width=\linewidth]{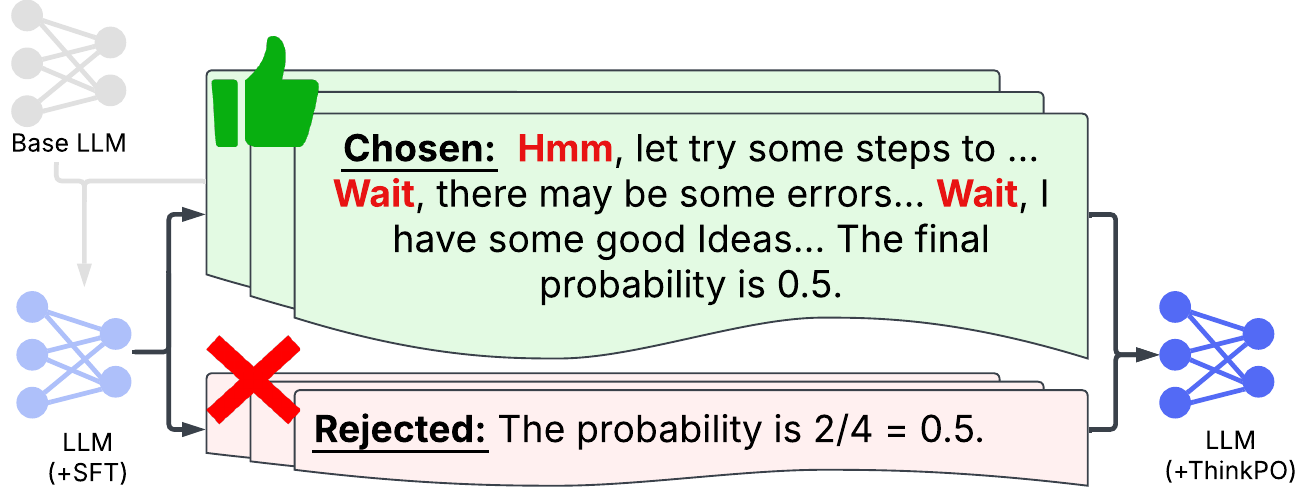}\vspace{8pt}
\hfill 
\includegraphics[height=0.980in]{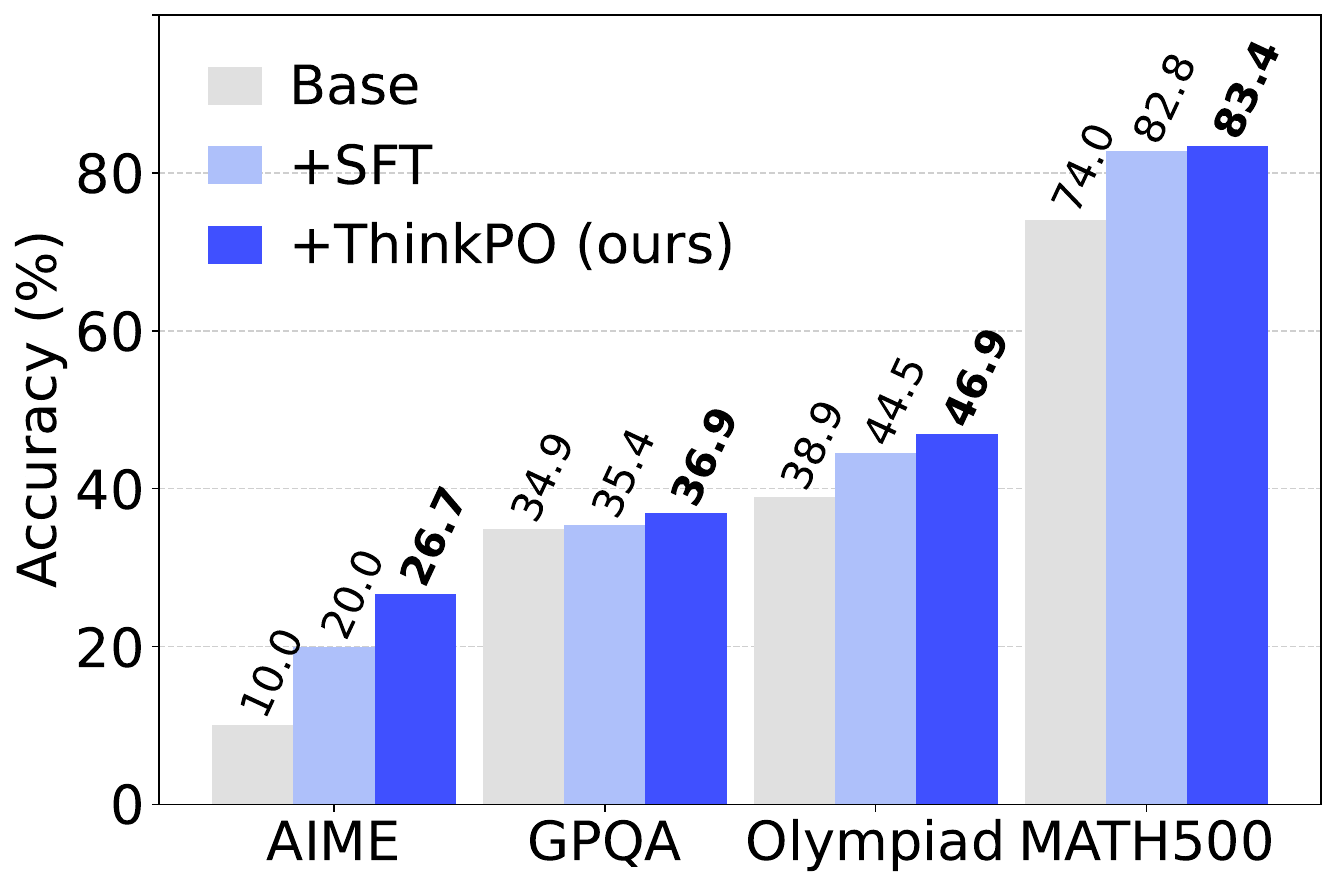}
\hfill
\includegraphics[height=0.990in]{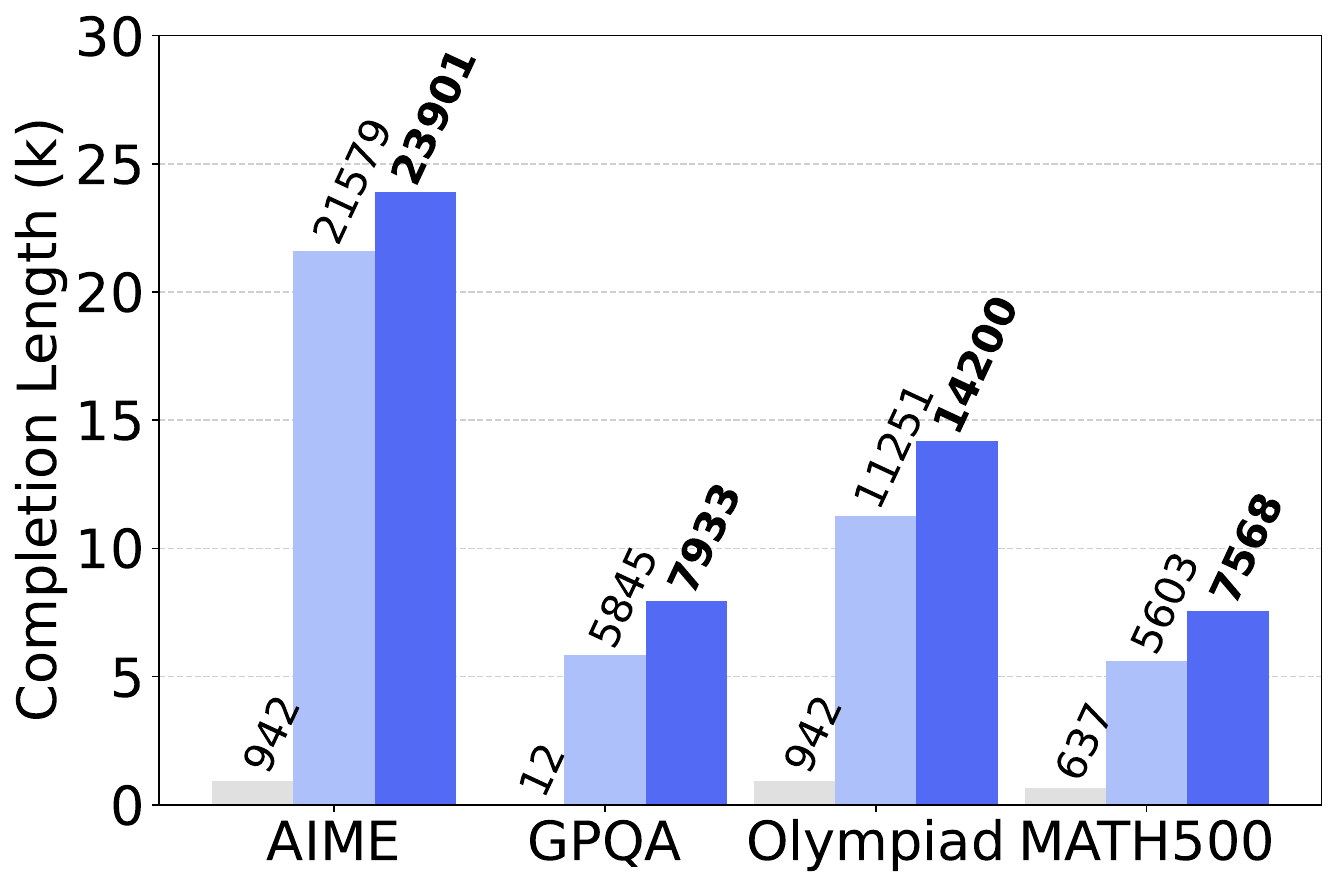}\vspace{-5pt}
  \caption{The illustration of our method ThinkPO and its performance on math reasoning tasks.
\textbf{Top:} Our ThinkPO enhances fine-tuned LLMs (+SFT) by promoting detailed problem-solving---using long chain-of-thought reasoning answers as positive (chosen) samples and short chain-of-thought reasoning answers as negative (rejected) samples.
\textbf{Bottom Left:} ThinkPO significantly boosts performance across mathematical benchmarks (e.g., 83.4\% on MATH500 vs. 82.8\% for +SFT and 74.0\% for the Base model).
\textbf{Bottom Right:} ThinkPO generates more detailed solutions, with average completion lengths on AIME increasing from 0.94K to 21.57K to 23.9K tokens.
These results underscore Think Preference Optimization's effectiveness in fostering and enhancing advanced mathematical reasoning.
  }
  \label{fig:open_source_lengths}
  \label{fig:Training pipeline}\vspace{-10pt}
\end{figure}

\section{Introduction}

\begin{figure*}[t]
\includegraphics[width=0.33\linewidth]{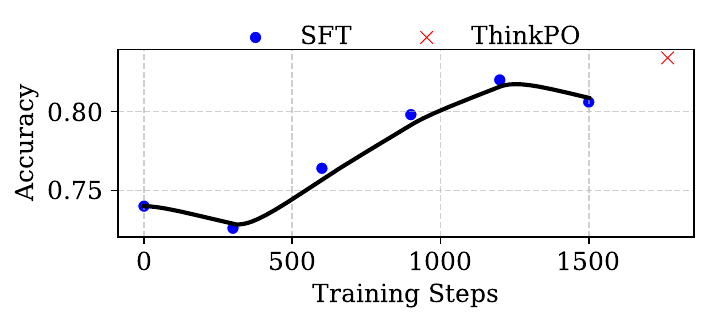}
\includegraphics[width=0.33\linewidth]{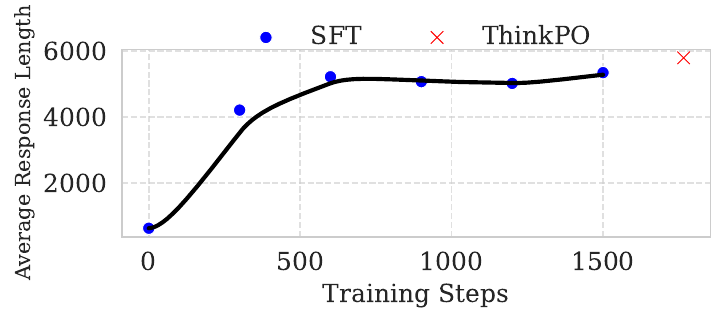}
\includegraphics[width=0.33\linewidth]{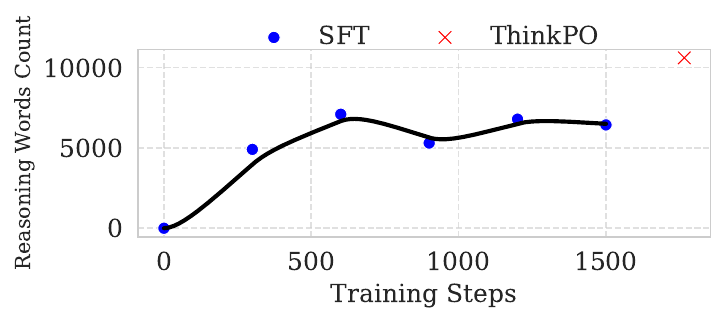}
\vspace{-20pt}
  \caption{
   Analysis of accuracy(\textbf{Left}), average response length(\textbf{Middle}) and reasoning-supportive words count(\textbf{Right}, like wait, hmm, etc) in SFT and ThinkPO process. We evaluate the model on MATH500 every 300 steps and record all the three metrics. In the early training stages, all of them improve significantly. However, in the later stages (e.g., after 1200 steps), the model’s performance gradually plateau. When ThinkPO is applied, we see additional improvements in all of the three aspects, demonstrating the effectiveness of Thinking Preference Optimization.
  }
  \label{fig:sft_length_plot}
\end{figure*}

 The reasoning capability of LLMs is crucial for their applicability in complex problem-solving tasks. Improving the reasoning ability of large language models is one of the current research hotspots. Many approaches have emerged in the open-source community that enhance relatively small models' reasoning ability through \textbf{SFT}. For example, Sky-Thought \cite{schulman2017proximal}, Bespoke-Stratos \cite{bespoke_stratos} and OpenThinker-7B\cite{openthoughts} have built long reasoning datasets to fine-tune models fully, aiming to improve model reasoning capabilities. Further advancements can be seen in models like s1 \cite{muennighoff2025s1simpletesttimescaling} and LIMO \cite{ye2025limoreasoning}, which focus on the sophisticated design of long reasoning datasets to enhancereasoning capabilities.

Despite the success of supervised fine-tuning, continually improving the reasoning abilities of the STF-ed model faces the following challenges:
(1) \textbf{high resources cost needed to collect new long reasoning response}: Training a stronger reasoning model first requires collecting new large-scale, diverse, and meticulously designed long-reasoning questions. Then, responses to these long reasoning problems need to be collected from large-scale models, such as DeepSeek-R1. However, collecting questions and responses requires significant computational power and human resources, making the process both expensive and labor-intensive. Furthermore,
(2) \textbf{repeatedly fine-tuning LLMs on existing long responses face Performance bottleneck}: As a compromise, one might repeatedly train on a limited long reasoning dataset, but this approach typically leads to a performance plateau or even decline. 
In \cref{fig:sft_length_plot}, we observe that when training with a fixed amount of long-reasoning data for multiple epochs, model’s average output length and accuracy increase significantly in the early stages but slow down or even plateau in later stages. According to the test-time scaling principle \cite{snell2024scaling, welleck2024decoding}, increasing the compute at test time generally enhances reasoning ability. However, the limited long-reasoning dataset is insufficient to further improve LLMs' reasoning capability in later stages of SFT.

To overcome the performance bottleneck and better utilize existing long reasoning data, we propose \textbf{Thinking Preference Optimization}: a simple yet efficient method to further enhance model reasoning ability after supervised fine-tuning (SFT). Our approach utilizes short CoT reasoning responses---which are already available or easy to acquire---as rejected answers and \textit{existing} long CoT responses as chosen answers for the same question, and employs Direct Preference Optimization to train models. This encourages models to prefer longer and more structured reasoning processes, thereby improving reasoning abilities {\textit{without acquiring additional high-quality long CoT responses}}.

\cref{fig:open_source_lengths} presents the framework of \TPO along with the experimental results. We first fine-tune a Qwen base model using the long CoT data to obtain an SFT-ed model (+SFT), and then we further train it using \TPO~ (+\TPO). The results in \cref{fig:open_source_lengths} clearly show that our method improves mathematical reasoning ability across four datasets. Additionally, our method increases the average response length on all four datasets, aligning with the test-time scaling trend. For example, ThinkPO increases the math reasoning accuracy of SFT-ed models by $8.6\%$ and the output length by $25.9\%$. Notably, \TPO~ increases the official DeepSeek-R1-Distill-Qwen-7B's performance on MATH500 from $87.4\%$ to $91.2\%$. The main contributions are summarized as follows:

\begin{itemize}[leftmargin=0.4cm, itemindent=.0cm, itemsep=0.0cm, topsep=0.1cm]
\item We propose Thinking Preference Optimization (\TPO) to maximize the value of existing long reasoning data, which successfully further enhances SFT-ed LLMs' reasoning performance without additional long CoT responses. 
\item Our method continuously improves the performance of public R1-distilled models, including the DeepSeek-R1 official distilled models. 
\item We release our dataset, codes, and model weights to facilitate further research. 
\end{itemize}
\section{Thinking Preference Optimization}

\subsection{Motivations}

This section introduces the motivations behind Thinking Prference Optimization. SFT with fixed long-reasoning datasets is an effective method for enhancing a model’s reasoning ability. However, further improvement of the model’s reasoning ability during the later stages faces a bottleneck. In such cases, by using short reasoning data as rejected samples and long reasoning texts from SFT as chosen samples for DPO training, it is possible to further leverage the high-quality SFT reasoning data to boost the model’s reasoning performance with minimal additional data resources.

First, we finetune Qwen-2.5-7B-Instruct model using Bespoke-Strato-dataset\cite{bespoke_stratos}, which includes $17k$ long reasoning data distilled from Deepseek-R1. During training, we track the model’s average output length, accuracy and reasoning-supportive words count (like wait, hmm) at different steps on the Math500 dataset. These are visualized by fitting curves. When calculating the model’s average output length, we only considered valid sentences, excluding duplicates or sentences with formatting errors. The results on other datasets could be found in \cref{Analysis of our Reproduce Model in other datasets}.

In \cref{fig:sft_length_plot}, 
in the early stages of SFT, the model’s average output length, accuracy and reasoning-supportive words count show significant improvements. This aligns with the test-time scaling phenomenon \cite{snell2024scaling, welleck2024decoding}, where a model’s reasoning ability generally improves as its output length increases. Many approaches enhance reasoning ability by fine-tuning models to generate longer responses. However, in the later stages of SFT, average response length, accuracy and reasoning-supportive words count plateau, indicating a performance bottleneck.

To further enhance the model’s reasoning ability, we can apply DPO, which encourages the model to favor longer outputs. By treating long-reasoning responses as chosen samples and short-reasoning responses as rejected samples, this approach improves the model’s reasoning ability without significantly increasing long-reasoning dataset size, thereby boosting its reasoning performance.

\subsection{Training Pipeline}

The training process in Thinking Preference Optimization consists of two stages: Reasoning SFT (Supervised Fine-Tuning) stage and Reasoning DPO (Direct Preference Optimization) stage.

In the Reasoning SFT stage, long-reasoning responses are collected for each question to construct the dataset $\mathcal{D}_{sft}$. The base model is then fine-tuned on $\mathcal{D}_{sft}$ to acquire advanced reasoning capabilities, which helps to prepare the model for next stage.

In the second stage, the model is further encouraged to generate extended reasoning using the Direct Preference Optimization (DPO) \cite{rafailov2024direct} approach. First, the long-reasoning data from the initial stage is used as the chosen responses. Then, a smaller model with normal Reasoning ability, such as Qwen-2.5-7B-Math \cite{qwen2.5}, is utilized to generate shorter reasoning responses as rejected samples. To ensure data quality, both long and short reasoning responses undergo filtering, including correctness validation. This process results in the dataset $\mathcal{D}_{dpo}$. Finally, the model trained in the first stage is fine-tuned on $\mathcal{D}_{dpo}$ using DPO, encouraging the model to generate longer outputs while enhancing its reasoning ability. Training pipeline is visualized as \cref{fig:Training pipeline}.

\begin{figure}[t]
\includegraphics[width=\linewidth]{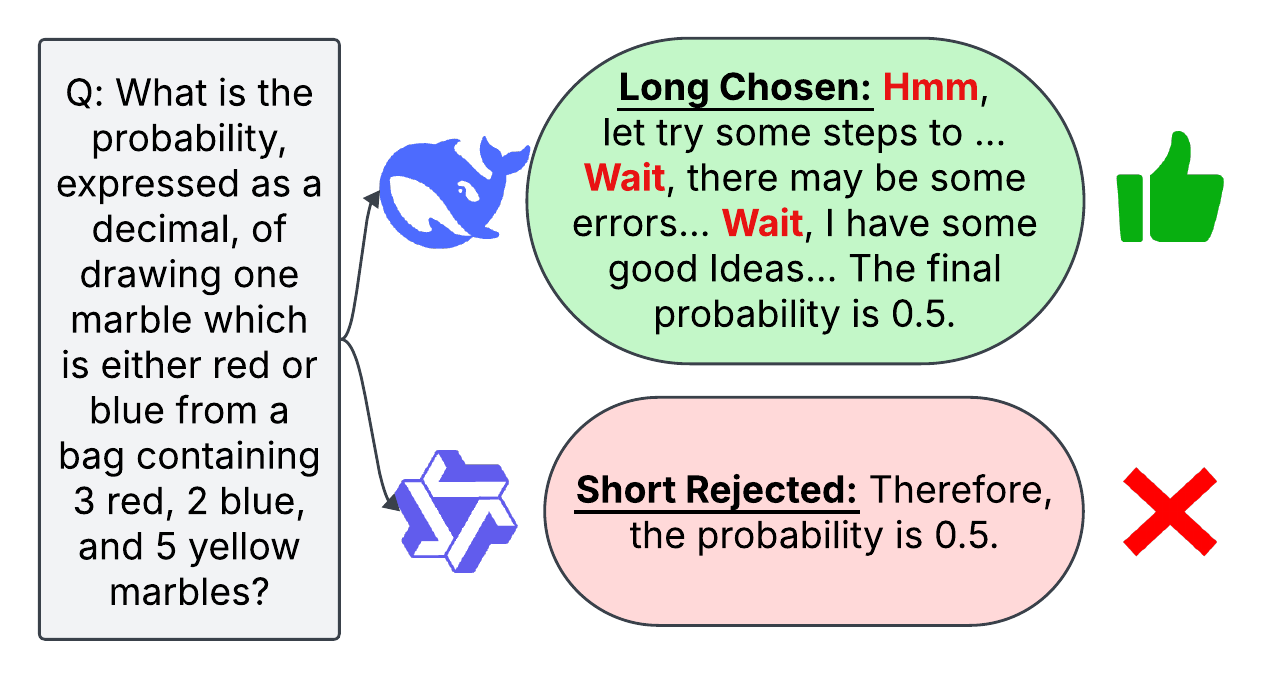}\vspace{-10pt}
  \caption{Data Collection Process: we use Deepseek R1 to generate long reasoning answers as chosen samples and Qwen 2.5-7B-Math to generate short reasoning answers as rejected samples, collecting datasets for DPO Training. Compare with short reasoning data, long reasoning answers includes many reasoning-supportive discourse markers, such as wait, hmm, and other hesitation cues, which can improve the model’s reasoning ability.}
  \label{fig:data curation}
\end{figure}
\begin{table*}[t]
\centering
\setlength{\tabcolsep}{8.5pt}
\caption{
  \label{table: Experiments on Bespoke-our}
  Accuracy and Average Response Length comparison for Our finetuned Qwen-2.5-7B-Instruct before and after
ThinkPO. The "Improv." column shows the percentage change of \textbf{Ours} over the model. After applying ThinkPO, its accuracy and length almost improve across datasets, further validating the effectiveness of \TPO.
}\vspace{-10pt}
\begin{tabular}{c|cccc|cccc}
\toprule
 \multicolumn{1}{c}{} & \multicolumn{4}{|c|}{Accuracy} & \multicolumn{4}{c}{Average Response Length}\\[0.5ex]
\midrule
 Dataset                        & {\small Base}       & {\small +SFT}        & {\small+\textbf{ThinkPO}} & \cellcolor{gray!20}{{\small Improv.(\%)}} 
                         & {\small Base}       & {\small +SFT}        & {\small +\textbf{ThinkPO}} & \cellcolor{gray!20}{{\small Improv.(\%)}}\\
\midrule
MATH500     & $74.0$ & $82.8$ & $\mathbf{83.4}$ & \cellcolor{gray!20}{\color{darkgreen}$0.7\%$} & $637$  & $5603$ & $\mathbf{7568}$  & \cellcolor{gray!20}{{\color{darkgreen}$35.0\%$}}  \\
AIME        & $10.0$ & $20.0$ & $\mathbf{26.7}$ & \cellcolor{gray!20}{{\color{darkgreen}$33.5\%$}}  & $942$ & $21579$ & $\mathbf{23901}$ & \cellcolor{gray!20}{{\color{darkgreen}$10.7\%$}}\\
GPQA        & $34.9$ & $35.4$ & $\mathbf{36.9}$ & \cellcolor{gray!20}{{\color{darkgreen}$4.2\%$}} & $12$ & $5845$ & $\mathbf{7933}$  & \cellcolor{gray!20}{{\color{darkgreen}$35.6\%$}}\\
GSM8K       & $90.1$ & $\mathbf{93.9}$ & $93.0$ & \cellcolor{gray!20}{$-0.9\%$}  & $260$ & $1310$ & $\mathbf{1599}$ & \cellcolor{gray!20}{{\color{darkgreen}$22.1\%$}}  \\
Olympiad    & $38.9$ & $44.5$ & $\mathbf{46.9}$ & \cellcolor{gray!20}{{\color{darkgreen}$5.4\%$}}  & $942$ & $11251$ & $\mathbf{14200}$ & \cellcolor{gray!20}{{\color{darkgreen}$26.2\%$}}\\
\midrule
Avg.        & $49.6$ & $55.3$ & $\mathbf{57.4} $     & \cellcolor{gray!20}{\color{darkgreen}$8.6\%$}  & $558$ & $9117$ & $\mathbf{11040}$   & \cellcolor{gray!20}{\color{darkgreen}25.9\%}\\
\bottomrule
\end{tabular}
\end{table*}
\begin{figure*}[t]
\includegraphics[width=0.49\columnwidth]{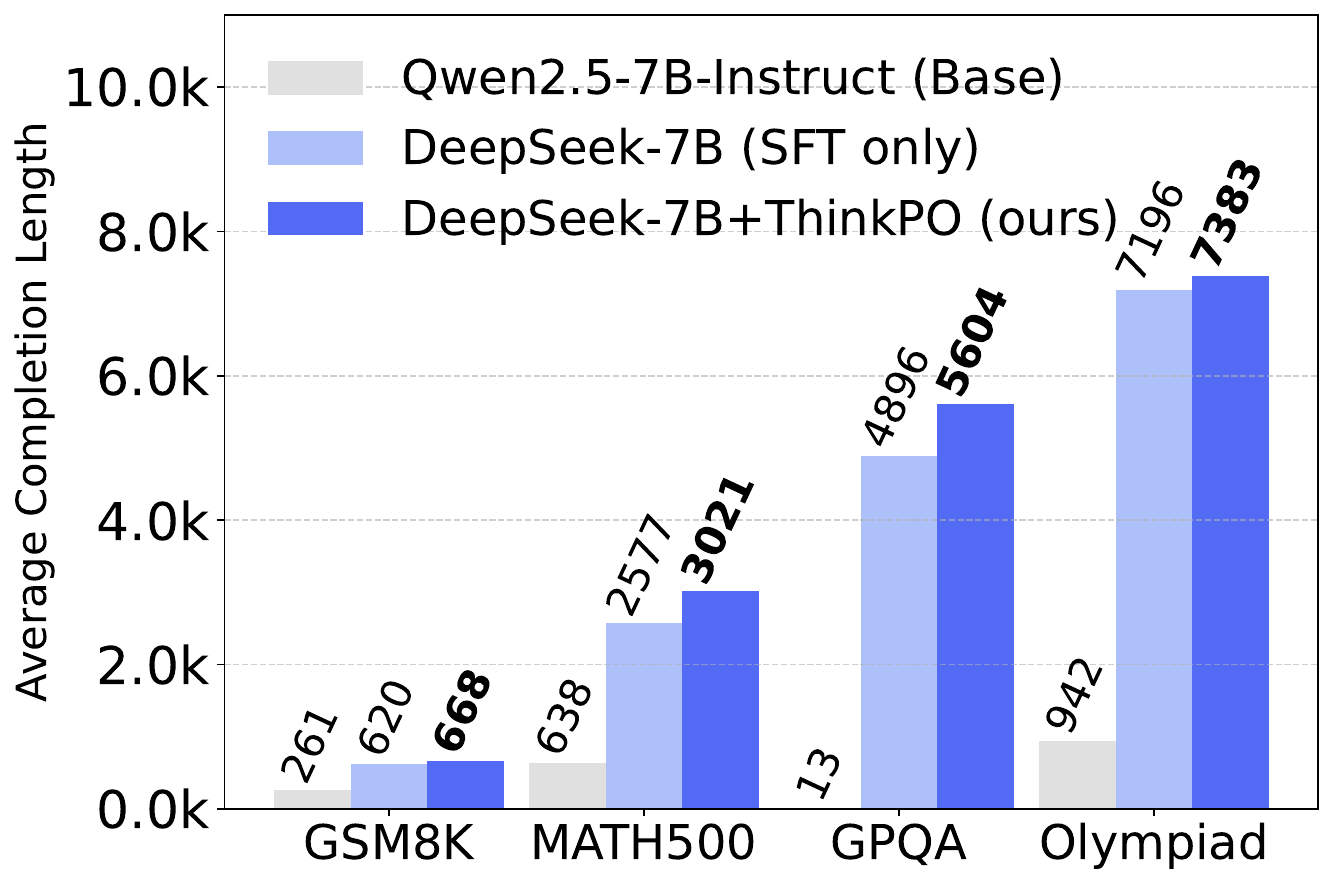}
\hfill
\includegraphics[width=0.49\columnwidth]{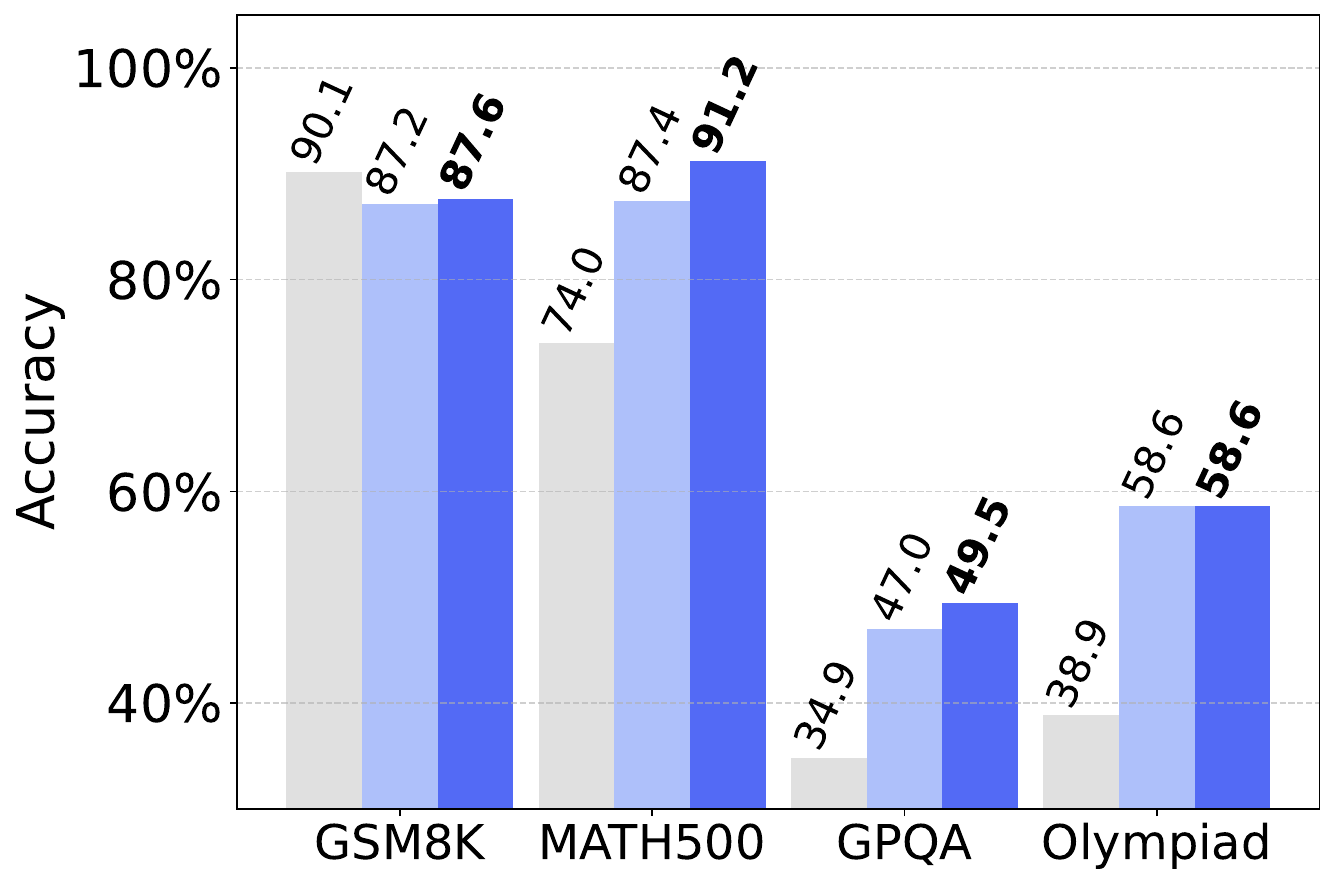}
\hfill
\includegraphics[width=0.49\columnwidth]{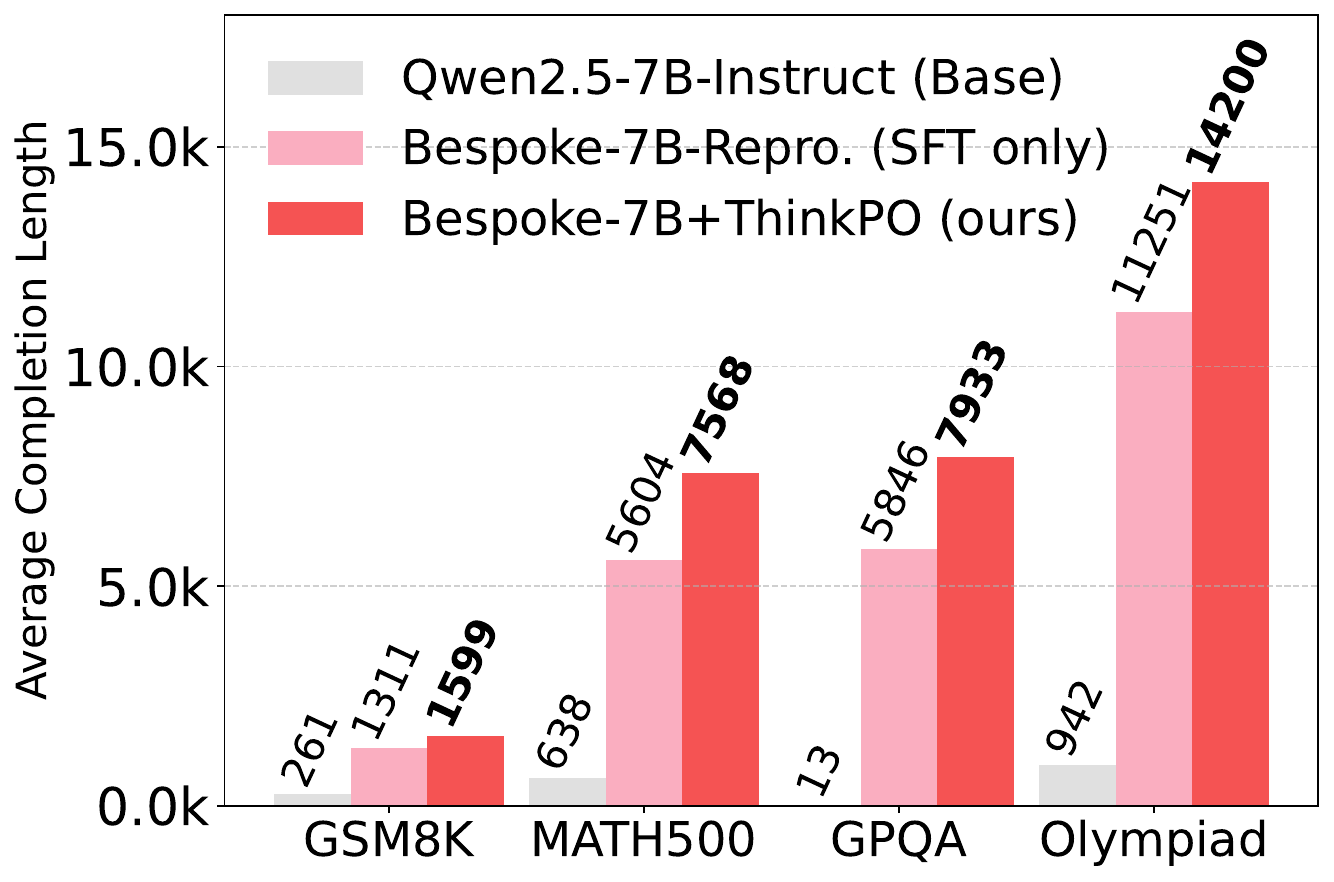}
\hfill
\includegraphics[width=0.49\columnwidth]{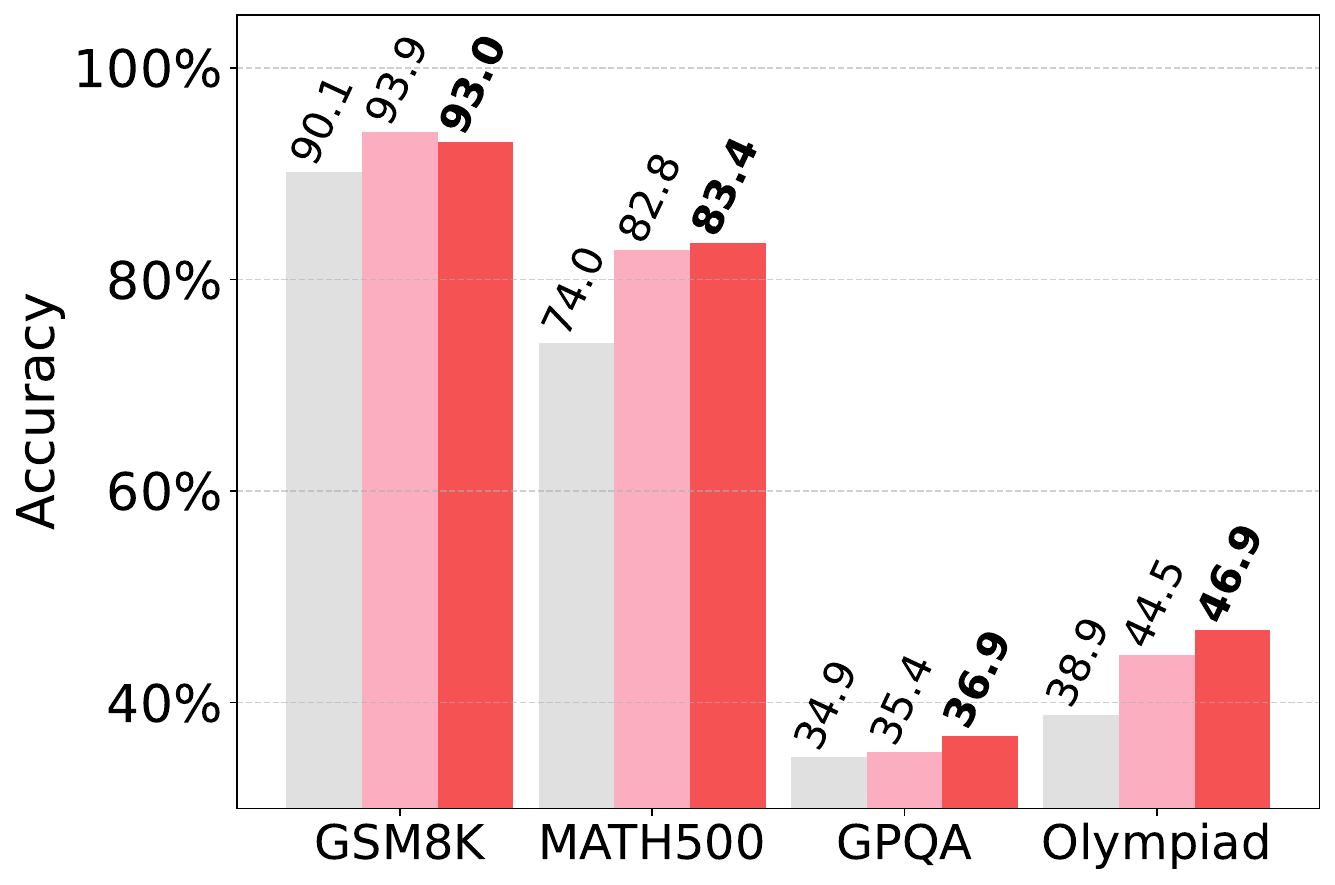}\vspace{-10pt}
\caption{Visualization of improvements on Accuracy and Average Response Length of DeepSeek-R1-Distill-Qwen-7B (\textbf{Left}) and our finetuned Qwen2.5-7B-Instruct (\textbf{Right}) on four datasets After ThinkPO. ThinkPO could improve DeepSeek-7B's and our finetuned Qwen2.5-7B's accuracy and output lengths almost across all the datasets }
  \label{fig: results of deepseek and repro}
\vspace{3pt}
\hfill
\includegraphics[width=0.32\linewidth]{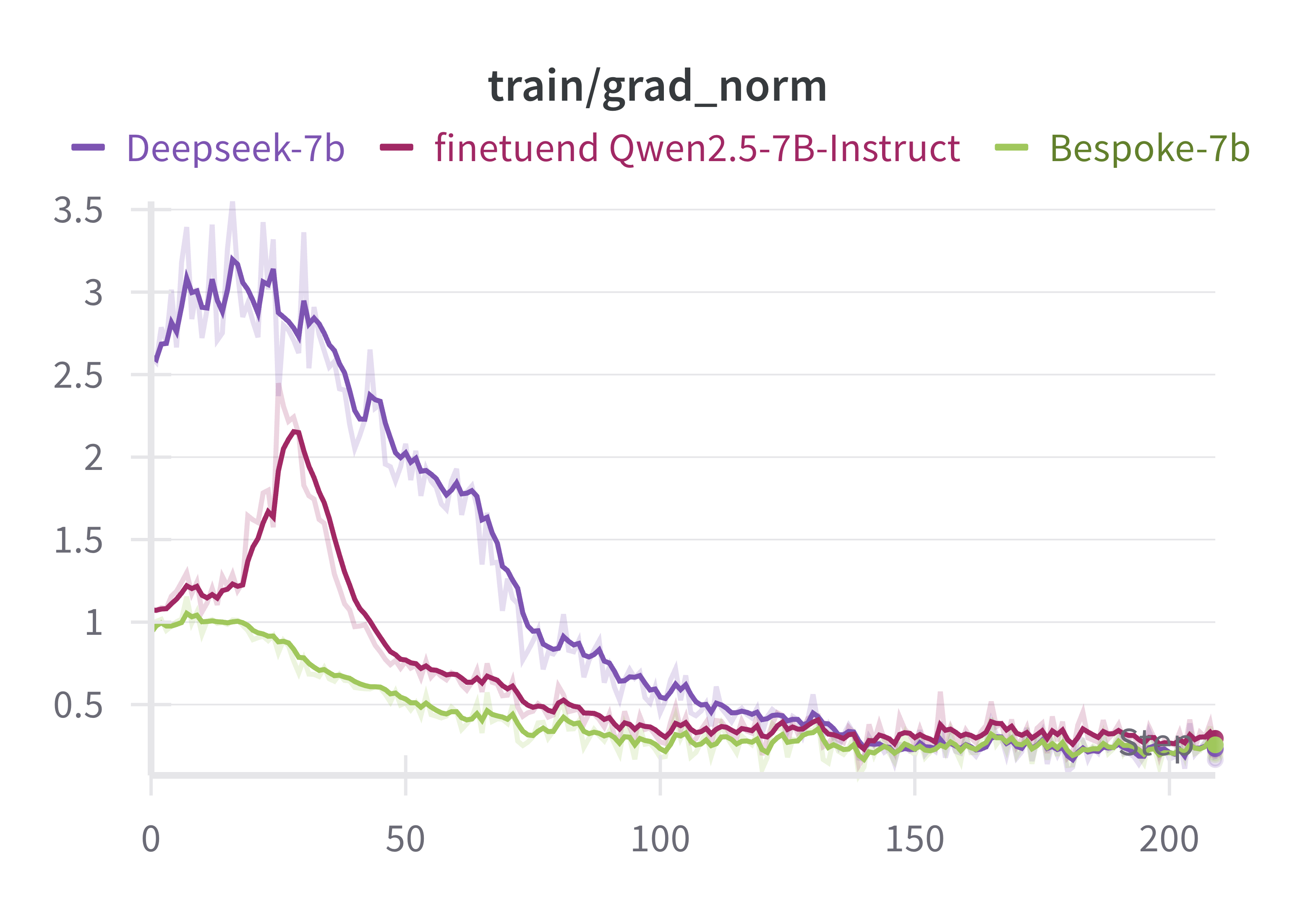}
\includegraphics[width=0.32\linewidth]{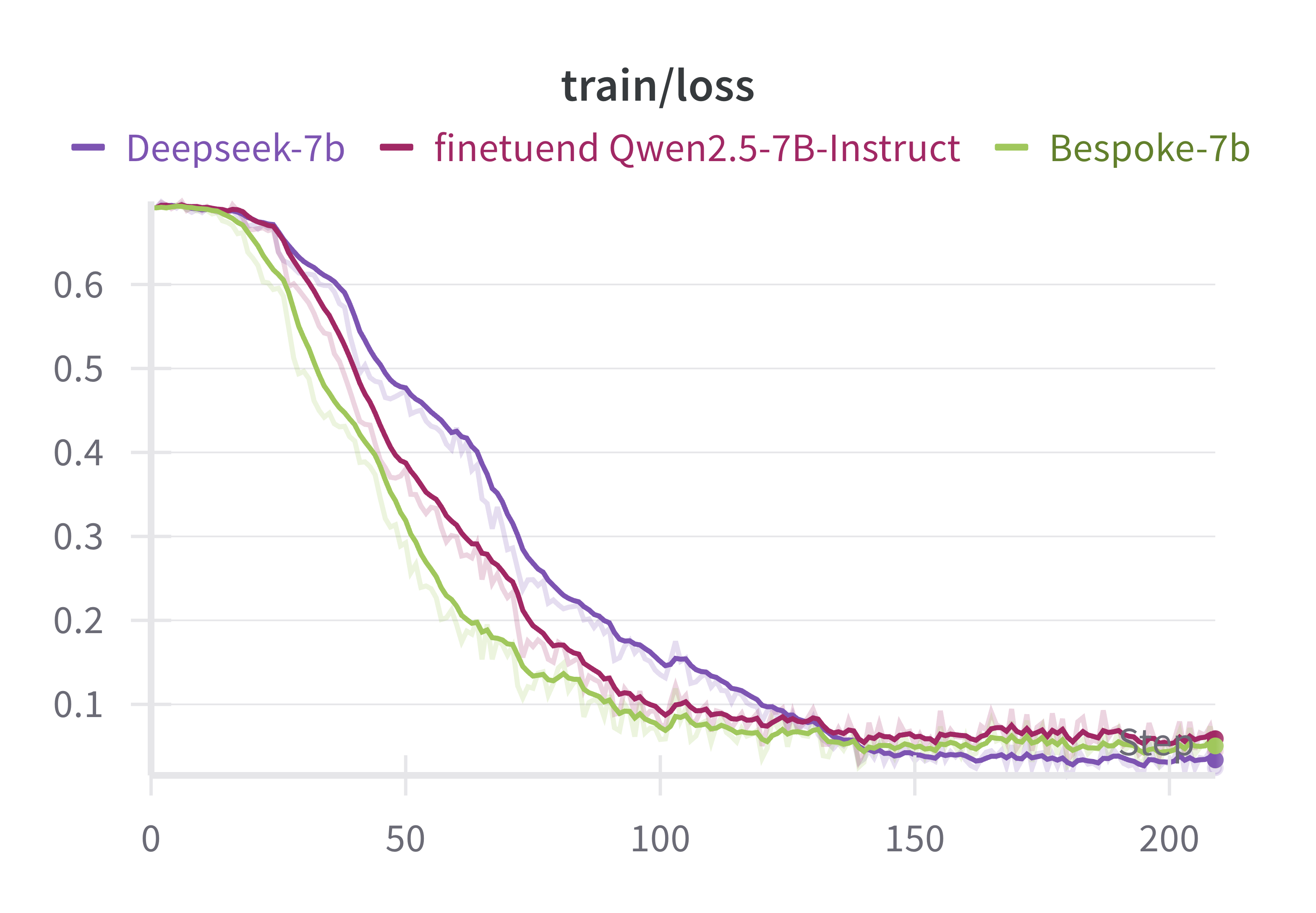}
\includegraphics[width=0.32\linewidth]{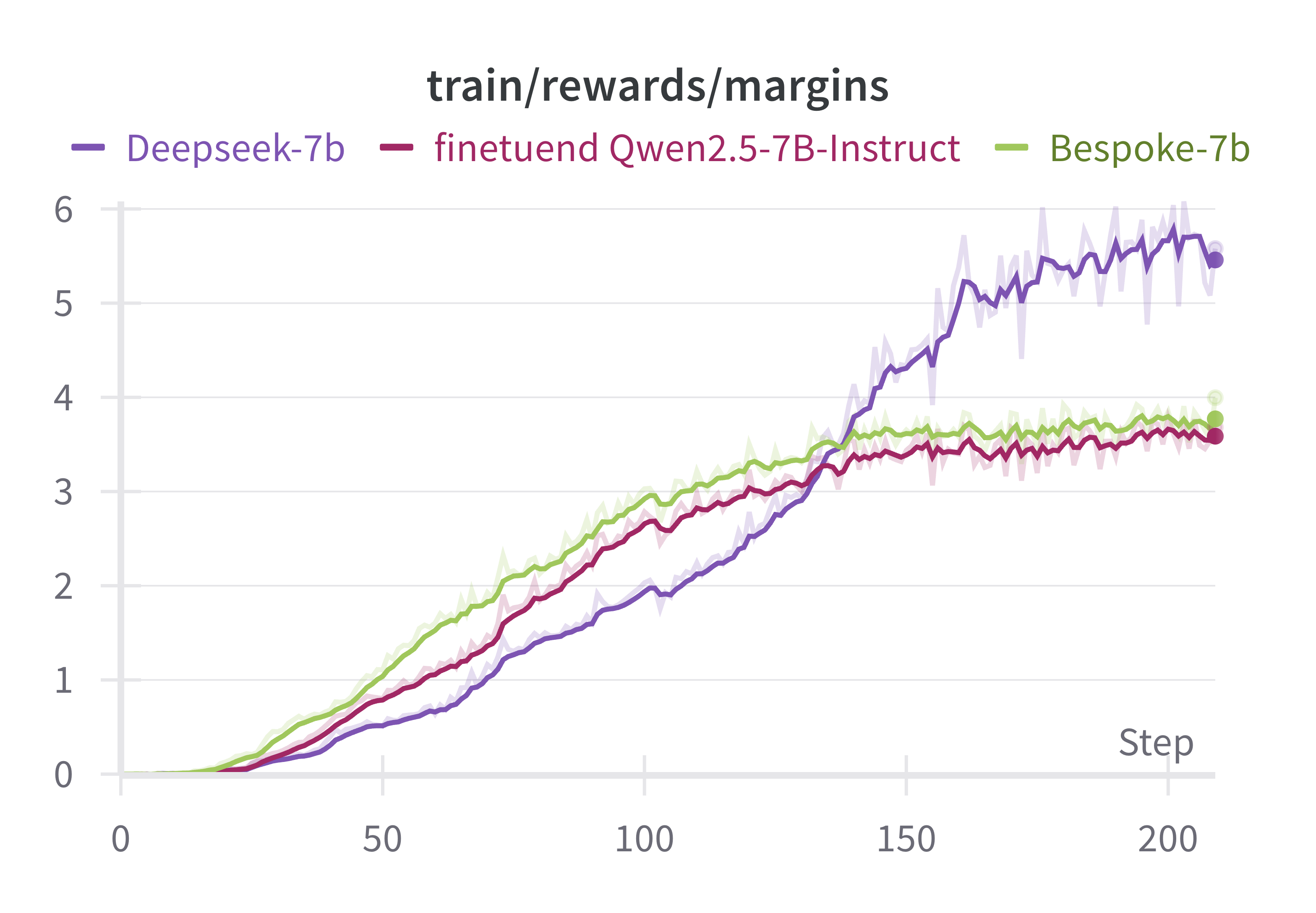}
\vspace{-10pt}
 \caption{
   Training loss, gradient norm, and margin curves for DeepSeek-R1-Distill-Qwen-7B, Bespoke-Stratos-7B and our finetued Qwen2.5-7B-Instruct during Thinking Preference Optimization phase.
  }
 \label{fig:dpo trainging loss}
\end{figure*}

\subsection{Data Curation}

The dataset $\mathcal{D}_{sft} = \{(q, o_{long})\}_{N}$ is based on bespoke stratos dataset \cite{bespoke_stratos}. They  used DeepSeek-R1 as the teacher reasoning model instead of QwQ-32B-Preview to generate long reasoning response $o_{long}$ and employed GPT-4o-mini in place of Sky-thought T1’s \cite{sky_t1_2025} parsing logic to filter out incorrect mathematical solutions.

For the dataset $\mathcal{D}_{dpo} = \{(q, o_{long}, o_{short})\}_{N}$ in the second stage, we collect it in the following manner, referring to \cite{kimiteam2025kimik15scalingreinforcement}: For each question $q$ in $\mathcal{D}_{sft}$, we use Qwen2.5-Math-7B-Instruct \cite{qwen2.5} to generate a short reasoning response~$o_{short}$~, pairing it with the long reasoning response~$o_{long}$ in $\mathcal{D}_{sft}$. We then retain the samples where Qwen2.5-Math-7B-Instruct’s answer matched DeepSeek R1’s answer, resulting in 8,080 samples. Additionally, we include 2,000 samples where Qwen2.5-Math-7B-Instruct’s answer differed from DeepSeek R1’s but adhered to the correct response format, including more output distribution in $\mathcal{D}_{dpo}$. All of these combined samples consequently form the final dataset $\mathcal{D}_{dpo}$. The dataset is collected through a straight foreword and simple process of gathering short-reasoning data, which did not require significant resources, compared to high-quality long-reasoning data. 


\begin{table*}[t]
\centering
\caption{Accuracy and Average Response Length comparison for Deepseek-7B and Bespoke-7B before and after ThinkPO. Qwen2.5-7B-Instruct shows the base performance, Deepseek-7B/Bespoke-7B report performance after SFT, and the "Improv." column shows the percentage change of \textbf{Ours} over Deepseek-7B/Bespoke-7B.}\vspace{-10pt}
\setlength{\tabcolsep}{12pt}
\begin{tabular}{c|ccc|ccc}
\toprule
\multicolumn{7}{c}{\href{https://huggingface.co/deepseek-ai/DeepSeek-R1-Distill-Qwen-7B}{DeepSeek-R1-Distill-Qwen-7B} (Deepseek)} \\[0.5ex]
\midrule
\multicolumn{1}{c}{} & \multicolumn{3}{|c|}{Accuracy} & \multicolumn{3}{c}{Average Response Length}\\[0.5ex]
\hline
\multirow{2}{*}{Dataset} & Deepseek   & \textbf{Ours}  & \cellcolor{gray!20}Improv.  & Deepseek   & \textbf{Ours}  & \cellcolor{gray!20}Improv. \\
                         & {\small(SFT)}        & {\small(+ThinkPO)} & \cellcolor{gray!20}{\small(\%)} 
                         & {\small(SFT)}        & {\small(+ThinkPO)} & \cellcolor{gray!20}{\small(\%)}\\
\hline
MATH500     & $87.4$ & $\mathbf{91.2}$ & \cellcolor{gray!20}{\color{darkgreen}$4.3\%$} & $2577$   & $\mathbf{3021}$ & \cellcolor{gray!20}{\color{darkgreen}$17.2\%$} \\
AIME        & $\mathbf{56.7}$ & $43.3$ & \cellcolor{gray!20}${-23.6\%}^{*}$ & $11419$  & $\mathbf{12875}$ & \cellcolor{gray!20}{\color{darkgreen}$12.8\%$} \\
GPQA        & $47.0$ & $\mathbf{49.5}$ & \cellcolor{gray!20}{\color{darkgreen}$5.3\%$}  & $4895$   & $\mathbf{5604}$ & \cellcolor{gray!20}{\color{darkgreen}$14.5\%$} \\
GSM8K       & $87.2$ & $\mathbf{87.6}$ & \cellcolor{gray!20}{\color{darkgreen}$0.5\%$}  & $619$    & $\mathbf{668}$  & \cellcolor{gray!20}{\color{darkgreen}$7.9\%$} \\
Olympiad    & $58.6$ & $58.6$       & \cellcolor{gray!20}$0.0\%$  & $7196$   & $\mathbf{7383}$ & \cellcolor{gray!20}{\color{darkgreen}$2.6\%$}  \\
\midrule
\multicolumn{7}{c}{\href{https://huggingface.co/bespokelabs/Bespoke-Stratos-7B}{Bespoke-Stratos-7B} (Bespoke)} \\[0.5ex]
\midrule
\multicolumn{1}{c}{} & \multicolumn{3}{|c|}{Accuracy} & \multicolumn{3}{c}{Average Response Length}\\[0.5ex]
\hline
\multirow{2}{*}{Dataset} & Bespoke    & \textbf{Ours}  & \cellcolor{gray!20}Improv.  & Bespoke    & \textbf{Ours}  & \cellcolor{gray!20}Improv. \\
                         & {\small(SFT)}        & {\small(+ThinkPO)} & \cellcolor{gray!20}{\small(\%)} 
                         & {\small(SFT)}        & {\small(+ThinkPO)} & \cellcolor{gray!20}{\small(\%)}\\
\hline
MATH500     & $\mathbf{84.0}$ & $82.8$ & \cellcolor{gray!20}$-1.4\%$  & $5696$   & $\mathbf{6404}$ & \cellcolor{gray!20}{\color{darkgreen}$12.4\%$} \\
AIME        & $20.0$ & $\mathbf{23.3}$ & \cellcolor{gray!20}{\color{darkgreen}$16.5\%$}  & $19858$  & $\mathbf{20079}$ & \cellcolor{gray!20}{\color{darkgreen}1.1\%} \\
GPQA        & $37.9$ & $\mathbf{43.4}$ & \cellcolor{gray!20}{\color{darkgreen}$14.5\%$}  & $5968$   & $\mathbf{7301}$ & \cellcolor{gray!20}{\color{darkgreen}$22.3\%$}\\
GSM8K       & $92.9$ & $\mathbf{93.3}$ & \cellcolor{gray!20}{\color{darkgreen}$0.4\%$} & $1404$   & $\mathbf{1755}$ & \cellcolor{gray!20}{\color{darkgreen}$25.0\%$}  \\
Olympiad    & $44.1$ & $\mathbf{48.5}$ & \cellcolor{gray!20}{\color{darkgreen}$10.0\%$} & $11140$  & $\mathbf{12204}$ & \cellcolor{gray!20}{\color{darkgreen}$9.6\%$}\\
\bottomrule
\end{tabular}
\parbox{\textwidth}{
        \scriptsize * Since AIME2024 contains only 30 questions, even a small difference in the number of correct answers can lead to significant fluctuations in accuracy, making the decline appear larger than it actually is.\\
    }
\vspace{-5mm}
\label{tab:accuracy_with_improvement}
\end{table*}

\section{Experiments}
\subsection{Experimental Setup}

To evaluate model’s reasoning ability, we select five different test sets: MATH500 \cite{lightman2023lets}, AIME2024~\footnote{\href{https://artofproblemsolving.com/wiki/index.php/AIME_Problems_and_Solutions}{AIME2024} is a math competition for high school students, acting as a qualifier for the USAMO.}, GPQA-Diamond \cite{rein2023gpqa}, GSM8K \cite{cobbe2021gsm8k}, and Olympiad Bench Math \cite{he2024olympiadbench}. These test sets primarily consist of mathematical reasoning problems, with GPQA-Diamond also including problems from physics, chemistry, and biology. The difficulty levels of these test sets vary significantly, with GSM8K being the easiest while AIME2024 is the most challenging. This diverse selection ensures a comprehensive assessment of the model’s reasoning capability across different levels of difficulty, from fundamental arithmetic to complex problem-solving with different difficulty. 

When generating responses, we set the temperature as 0.7. For results on other temperatures, please refer to \cref{Evaluating ThinkPO with Different Temperatures}. We present our chosen hyper-parameters of ThinkPO, such as learning rate, batch size and $\beta$, in \cref{Training Recipe}.

\subsection{Effectiveness of ThinkPO}

This experiment primarily analyzes the average response length, accuracy and reasoning-supportive words count during both SFT and DPO processes to validate the effectiveness of Thinking Preference Optimization (ThinkPO). By tracking these metrics, we aim to demonstrate how ThinkPO enhances the model’s reasoning ability by encouraging longer, more structured outputs, ultimately leading to improved reasoning performances.

First, we fine-tune Qwen-2.5-7B-Instruct with Bespoke-Stratos-Dataset. Subsequently, we apply \TPO~to enhance the model’s reasoning ability. The final results are shown in \cref{table: Experiments on Bespoke-our}. Our finetuned model achieves scores across the five datasets that are almost identical to  Bespoke-Stratos-7B, which is also finetuned on Bespoke-Stratos-Dataset, confirming the correctness of our SFT process. Furthermore, after applying \TPO, our model demonstrates improvements on almost all the datasets, validating the effectiveness of ThinkPO in enhancing and improving LLM reasoning ability.

Additionally, we analyze average response length and reasoning-supportive words (like \textit{wait}, \textit{hmm}, etc) at different steps during both SFT and ThinkPO. We record the model’s average response length, accuracy and reasoning-supportive words (like wait, hmm, etc) count on Math500 at different training steps, distinguishing between the SFT and \TPO. When calculating average response lengths, we exclude duplicate or incomplete responses to ensure accuracy. Additionally, when counting reasoning-supportive words, we only consider correct answers to prevent excessive occurrences of filler words like “wait” due to underthinking \cite{chen2024not, kirk2023understanding, wang2025thoughts}. The results are visualized in \cref{fig:sft_length_plot}.

At the initial stage of SFT, the model’s reasoning ability improves significantly. In the later stages of SFT (like after 1200 steps), three metrics gradually plateau, indicating that the model may have reached a local optimum. However, after applying Thinking Preference Optimization, model’s average response length, reasoning-supportive words count and accuracy improve, showing the effectiveness of ThinkPO in overcoming this stagnation. We visualize the trend of output length and accuracy across training steps on other datasets(like GSM8K). For more details, please refer to \cref{Analysis of our Reproduce Model in other datasets}.

\begin{table*}
  \centering
  \setlength{\tabcolsep}{5.5pt}
  \caption{
  \label{table: Experiments on different model sizes}
Results of Models with Different Sizes (3B, 7B, 14B) on the Qwen-2.5 Family. We evaluate models of different sizes (3B, 7B, 14B) trained with Supervised Fine-Tuning (SFT) and Think Preference Optimization (ThinkPO). Models are fine-tuned on the Bespoke-Strato-Dataset for 1 epoch.  As model size increases, accuracy improves across all five test datasets. After ThinkPO training, accuracy improves consistently for models of all sizes, including the smallest (3B), demonstrating that ThinkPO enhances reasoning ability across different model scales.
  }\vspace{-10pt}
  \begin{tabular}{c|ccc|ccc|ccc}
  \toprule
   & \multicolumn{3}{c|}{Qwen 2.5-3B} & \multicolumn{3}{c|}{Qwen 2.5-7B} & \multicolumn{3}{c}{Qwen 2.5-14B} \\
   & {\small+SFT} &  {\small+\textbf{\TPO}} & Improv. &  {\small+SFT} &  {\small+\textbf{\TPO}} & Improv. & {\small+SFT} &  {\small+\textbf{\TPO}} & Improv. \\
  \midrule
   {MATH500} & $53.6$ & $\mathbf{54.6} $ & \cellcolor{gray!20}{\color{darkgreen}$1.8\%$} & $73.0$ & $\mathbf{74.6}$  & \cellcolor{gray!20}{\color{darkgreen}$2.2\%$} & $83.2$ & {$\mathbf{85.6}$}  & \cellcolor{gray!20}{\color{darkgreen}$2.9\%$} \\
  {AIME}  &$3.30$ &  $\mathbf{6.7}$ & \cellcolor{gray!20}{\color{darkgreen}$100\%$ } & $\mathbf{16.7}$ & $13.3$  & \cellcolor{gray!20}${-20.3\%}^*$ & $23.3$ & $\mathbf{33.3}$ & \cellcolor{gray!20}{\color{darkgreen}$42.9\%$} \\
    {GPQA} & $26.3$ & $\mathbf{27.3}$ & \cellcolor{gray!20}{\color{darkgreen}$3.8\%$} & $32.3$ & $\mathbf{36.4}$ & \cellcolor{gray!20}{\color{darkgreen}$12.7\%$} & $\mathbf{45.5}$ &  $44.0$ & \cellcolor{gray!20}$-3.2\%$ \\
   {GSM8K} & $80.4$ &  $\mathbf{81.1}$ & \cellcolor{gray!20}{\color{darkgreen}$0.8\%$} &$88.2$ & $\mathbf{88.9}$ & \cellcolor{gray!20}{\color{darkgreen}0.9\%} & $93.7$ & $\mathbf{93.9}$& \cellcolor{gray!20}{\color{darkgreen}$0.2\%$} \\
   {Olympiad}& $20.0$ & $\mathbf{22.0}$ & \cellcolor{gray!20}{\color{darkgreen}$10.0\%$} & $35.3$ & $\mathbf{37.2}$ & \cellcolor{gray!20}{\color{darkgreen}$5.3\%$} & $49.9$ & $\mathbf{52.1}$& \cellcolor{gray!20}{\color{darkgreen}$4.4\%$ }\\
   \bottomrule
  \end{tabular}
  \parbox{\textwidth}{
        \scriptsize * Since AIME2024 contains only 30 questions, even a small difference in the number of correct answers can lead to significant fluctuations in accuracy, making the decline appear larger than it actually is.\\
    }
 \vspace{-5mm}
\end{table*}
\begin{figure*}[t]
\includegraphics[width=0.33\linewidth]{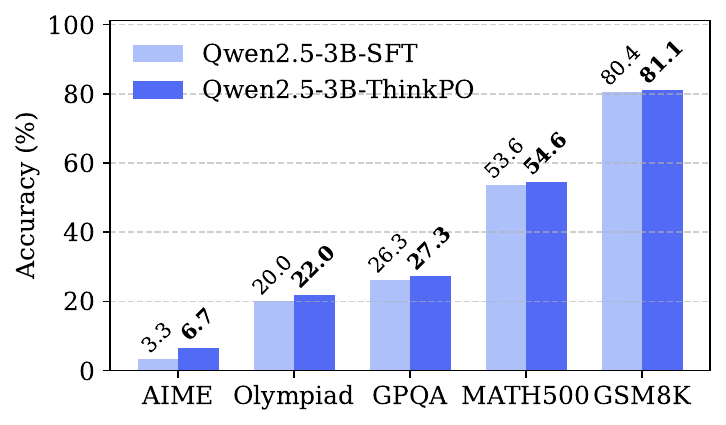}
\includegraphics[width=0.33\linewidth]{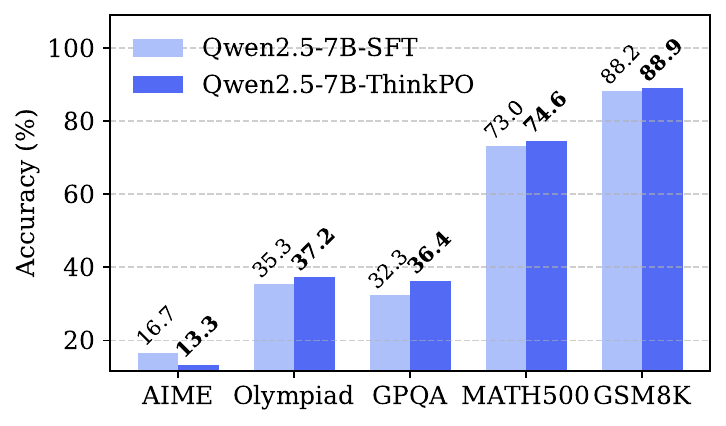}
\includegraphics[width=0.33\linewidth]{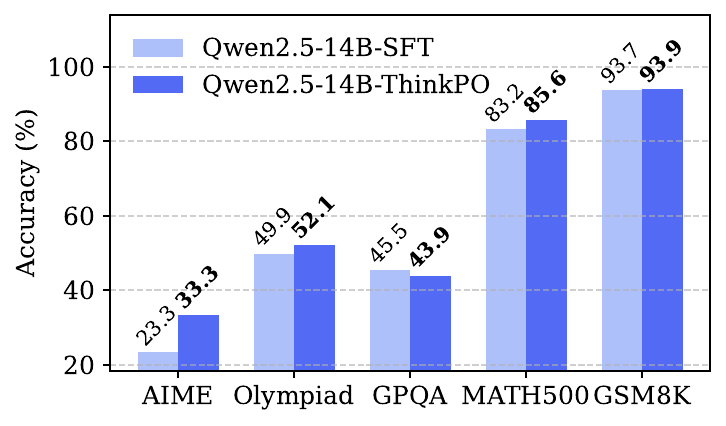}
\vspace{-25pt}
  \caption{Visualization of improvements on Accuracy and Average Response Length of models in the same family series from different sizes (Qwen-2.5-3B, Qwen-2.5-7B and Qwen-2.5-14B) on five datasets after ThinkPO. ThinkPO could improve models' accuracy and output lengths almost across all the datasets, regradless of sizes }
  \label{fig: TPO on different size}
\end{figure*}
\subsection{ThinkPO can Continually Improve Reasoning Ability of Public Distilled Models}

We select two open-source reasoning models and perform ThinkPO training using $\mathcal{D}_{dpo}$. Specifically, we chose \href{https://huggingface.co/deepseek-ai/DeepSeek-R1-Distill-Qwen-7B}{DeepSeek-R1-Distill-Qwen-7B} and \href{https://huggingface.co/bespokelabs/Bespoke-Stratos-7B}{Bespoke-Stratos-7B}, since both reasoning models were fine-tuned on Qwen2.5-7B-Instruct. 

As shown in \cref{tab:accuracy_with_improvement} and \cref{fig: results of deepseek and repro}, both models demonstrate an improvement in accuracy across five datasets. For example, Bespoke-Stratos-7B shows an increase in accuracy on all datasets except for a slight decline on the MATH500 dataset. Notably, the improvements on Olympiad Bench Math and GPQA-Diamond reach around $5\%$. DeepSeek-R1-Distill-Qwen-7B, with the exception of a decline on AIME2024, shows consistent or slightly improved accuracy. Specifically, on MATH500, the accuracy improves from 87.4\% to 91.2\%. 

In addition to accuracy, average response length of DeepSeek-R1-Distill-Qwen-7B is increased by around 500 tokens on the MATH500 dataset, while Bespoke-Stratos-7B shows a larger increase of approximately 1000 tokens. These align with test-time scaling principle \cite{snell2024scaling, welleck2024decoding}, where the increased response length reflects an enhancement in reasoning capacities.

\subsection{ThinkPO Works for Different-Size Models}

Previous experiments are all conducted using a 7B model for training. Now we utilize the Bespoke Stratos dataset and conduct one epoch of SFT training on models of varying sizes within the Qwen2.5 series (Qwen2.5-3B, Qwen2.5-7B, and Qwen2.5-14B). The learning rate is set to 3e-5, and other hyperparameters are kept consistent with Bespoke-Stratos-7B, ensuring the models' performances. The results after SFT and \TPO~ are presented in \cref{table: Experiments on different model sizes} and \cref{fig: TPO on different size}. First, as the model scale increases, its accuracy improves across all the datasets after SFT, which aligns with expectations. After applying \TPO, all models, regardless of size, achieve further improvements. Specifically, on Math500, all three models show an accuracy increase of 1\%–2\%. After applying ThinkPO, the Qwen2.5-3B model achieves accuracy improvements across all five datasets, while Qwen2.5-7B and 14B models show improvements on four datasets, which shows that ThinkPO is effective across different model scales, further validating its generalizability and robustness.

\section{Ablation}

\begin{table}
  \centering
  \caption{
  \label{table: Experiments on short data}
Results of ThinkPO on the model finetuned with a short-Reasoning Dataset. We select a short-chain reasoning dataset of the same size as the Bespoke-Stratos dataset and fine-tune Qwen-2.5-7B for 3 epochs. Models trained with reasoning-style datasets, regardless of response length, can benefit from \TPO~to enhance and improve their reasoning capability 
  }\vspace{-10pt}
  \begin{tabular}{c|ccc}
  \toprule
   \multicolumn{1}{c}{} & 
   Short & \textbf{Our} & Improv.  \\
   \multicolumn{1}{c}{} & {\small+SFT} &  {\small+\textbf{\TPO}} & \% \\
  \midrule
   \textbf{MATH500} & $57.8$ & $\mathbf{59.0}$ & \cellcolor{gray!20}{\color{darkgreen}$2.4\%$} \\
  \textbf{AIME}  &$0.0$ & $\mathbf{3.3}$ & \cellcolor{gray!20}{\color{darkgreen}$100\%$} \\
    \textbf{GPQA} & $30.3$ & $\mathbf{31.3}$ & \cellcolor{gray!20}{\color{darkgreen}$3.3\%$} \\
   \textbf{GSM8K} &$83.4$ & $\mathbf{85.1}$& \cellcolor{gray!20}{\color{darkgreen}$2.0\%$} \\
   \textbf{Olympiad}& $23.3$ & $\mathbf{23.6}$& \cellcolor{gray!20}{\color{darkgreen}$1.2\%$} \\
    \bottomrule
  \end{tabular}
\end{table}

\subsection{Whether ThinkPO is Useful when SFT with Short Reasoning Data?}

In our previous experiments, we fully fine-tuned the model using long reasoning datasets before applying ThinkPO to further enhance its reasoning ability. However, an important question arises: If we use short reasoning data instead of long reasoning data during the full fine-tuning stage, can Thinking Preference Optimization still improve the model’s reasoning performance effectively?

To investigate this, we conduct the following experiment. We use Qwen2.5-7B as the base model and select a dataset from AI-MO/NuminaMath-CoT\cite{numina_math_datasets} that matches the Bespoke-Stratos dataset with the same data size for fine-tuning. Unlike our previous experiments, the fine-tuning data here consists of short reasoning examples rather than long reasoning ones. Consequently, the fine-tuned model is expected to underperform compared to models trained on long-reasoning data. To equip models with basic reasoning ability, we fine-tune them for three epochs and set learning rate as 1e-5. Following this, we apply Thinking Preference Optimization using the same dataset in the previous experiments, aiming to further enhance and improve the model’s reasoning performance. 

As shown in \cref{table: Experiments on short data}, even after fine-tuning on short-reasoning data, ThinkPO still effectively improves the model’s reasoning ability. For example, on the Math500 dataset, after applying ThinkPO, the model’s accuracy improves by approximately 2\%. This result demonstrates that models trained with reasoning-style datasets, regardless of response length, can benefit from ThinkPO to enhance and improve their reasoning capability.

\subsection{Exploring the Impact of Length Differences between Chosen and Rejected Samples on ThinkPO.}

\begin{figure}[t]
\includegraphics[width=\columnwidth]{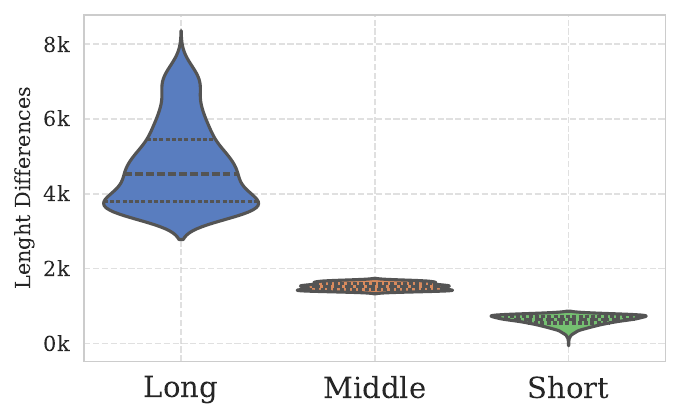}\vspace{-10pt}
  \caption{
    Length difference distribution between chosen and rejected samples across three datasets. These three datasets are 1000 samples selected based on the length difference from our \TPO-Dataset. The long dataset exhibits the widest distribution of length differences, while the middle and short datasets have smaller differences with lower mean values and variances.   
  }
  \label{fig:length difference}
\end{figure}
\begin{table}
  \centering
  \caption{
  \label{table: Experiments on different length}
   Model performance across three datasets with varying chosen and rejected sample length difference distributions. “Avg Differences” represents the average length difference between chosen and rejected samples. \textit{Short} yields the best overall performance, suggesting that appropriate length differences improve ThinkPO learning, while too large differences may hinder it.
  }\vspace{-10pt}
  \begin{tabular}{c|ccc}
  \toprule
   \multicolumn{1}{c}{} & Short & Middle & Long \\
   \midrule
   \textbf{Avg Differences} & $621$ & $1525$ & $4758$ \\
  \midrule
   \textbf{MATH500} & $\mathbf{84.2}$ & $81.8$ & $\mathbf{84.0}$ \\
  \textbf{AIME}  & $\mathbf{26.7}$ &$ 13.3$ & $16.7$\\
    \textbf{GPQA} &$ 40.9$ & $\mathbf{41.9}$ & $38.9$\\
   \textbf{GSM8K} &$92.9 $& $92.9$ & $\mathbf{93.0}$\\
   \textbf{Olympiad}& $\mathbf{46.1}$ & $45.9$& $45.9$\\
    \bottomrule
  \end{tabular}
\end{table}

In the entire ThinkPO dataset, we select long reasoning data as chosen and short reasoning data as rejected. A key question is whether the length disparity between chosen and rejected samples affects the ThinkPO training because length disparity is not distributed evenly in the dataset. To investigate this, we conduct an experiment to verify the impact of length differences on the ThinkPO training.

The ThinkPO dataset contains approximately 10,000 samples, but the length disparity between chosen and rejected samples is not uniformly distributed. Therefore, we select three datasets with different length distributions: short, middle, and long, each containing 1,000 samples. \cref{fig:length difference} shows details of the length differences distributions between chosen and rejected samples in these three datasets, with the long dataset exhibiting the largest and most widely distributed differences, the middle dataset showing moderate differences, and the short dataset having the smallest differences.

\cref{table: Experiments on different length} displays the results after ThinkPO for one epoch, using the Bespoke-Stratos-7B model as the base model. Each dataset shows certain advantages across the five test datasets. However, the short dataset yields the best performance on overall datasets. We propose that when the length difference is smaller, the model’s output distributions for both samples are more consistent, which benefits ThinkPO learning. On the other hand, when it is too large, it may not help the model’s learning.

\section{Related Works}

\textbf{LLM Reasoning Ability}. With the development of large models, reasoning ability \cite{wang2022self,zhang2023multimodal,yao2023tree,plaat2024reasoning} has become one of the most crucial capabilities and a necessary condition for achieving AGI (Artificial General Intelligence) 
\cite{minaee2024large,xu2024survey,morris2023levels,feng2024far,krishnan2025artificial}. The earliest appearance of long-chain reasoning ability in large models can be traced to OpenAI o1 \cite{jaech2024openai, arrieta2025o3,hurst2024gpt}, which excelled across various mathematical reasoning test sets and outperform contemporary LLMs.

This was followed by the release of the QwQ model \cite{qwen2.5, bai2023qwen, bai2023qwens, chu2024qwen2}, which trained reasoning capabilities using a process reward model approach \cite{li2024process, ma2023let, zhang2025lessons, lambert2024rewardbench}. Currently, the emergence of DeepSeek R1 \cite{deepseekai2025deepseekr1incentivizingreasoningcapability} and Kimi 1.5 \cite{kimiteam2025kimik15scalingreinforcement} has further enhanced the reasoning abilities of large open-source models. DeepSeek R1 utilizes a simple rule-based reward model \cite{ramesh2024group,hu2025reinforce++, shao2024deepseekmath, alonso2025mathematics, kirk2023understanding, yang2024bayesian} to effectively boost the model’s reasoning performance, bringing about an aha moment that narrows the reasoning capability gap between open-source and closed-source models. On the other hand, Kimi 1.5 employs several tricks, such as long-to-short reasoning, to achieve high efficiency in LLM reasoning performance.

Many works on open-source reasoning models have also emerged. First is Sky-Thought T1 \cite{sky_t1_2025}, which uses QwQ-32B-Preview as a teacher model to generate reasoning answers for training data. Then, Bespoke-Stratos \cite{bespoke_stratos} built upon Sky-Thought T1, using DeepSeek R1 as the teacher model to generate answers for Sky-Thought data. Since DeepSeek R1 has far superior reasoning abilities compared to QwQ-32B-Preview, the generated data quality is higher, allowing Bespoke-Stratos-7B and Bespoke-Stratos-32B models to achieve DeepSeek-level advanced reasoning performance after training on around 17k data points. Recently, s1 \cite{muennighoff2025s1simpletesttimescaling} and LIMO \cite{ye2025limoreasoning} have emphasized that fine-tuned, high-quality data construction is essential for models to achieve SOTA reasoning capabilities. 

\textbf{Direct Preference Optimization}. RLHF 
 \cite{chaudhari2024rlhf, kirk2023understanding, kaufmann2023survey} is designed to align model outputs with human preferences after supervised fine-tuning (SFT). Various methods have been introduced, such as Proximal Policy Optimization (PPO) 
 \cite{engstrom2019implementation, huang2022a2c, wijmans2019dd}. However, PPO is an online method that requires significant computational resources. To address this, Direct Preference Optimization was proposed, enabling offline training with only chosen and rejected sample pairs while reducing computational costs compared to PPO. Recently, several DPO variants \cite{wu2024beta,wu2024alpha, qi2024online, zhong2024dpo,su2025reveal} have emerged, including StepDPO \cite{lai2024step}, KTO \cite{ethayarajh2024kto}, SimPO \cite{meng2024simpo}, LongDPO \cite{ping2025longdpo}, Test-Time Preference Optimization \cite{li2025testtimepreferenceoptimizationonthefly} etc. Among them, LongDPO shares similarities with our proposed method. However, LongDPO primarily focuses on improving long-form story generation instead of reasoning abilities.

\section{Conclusion}

We introduce Thinking Preference Optimization, a simple yet effective post-SFT method without the need for additional high-quality long-reasoning data. By leveraging short reasoning responses as rejected and long reasoning responses as chosen, \TPO~encourages models to generate detailed reasoning outputs, effectively maximizing the utility of existing long-reasoning data. Our experiments demonstrate that \TPO~significantly improves model performance, yielding an 8.6\% accuracy boost and a 25.9\% increase in output length for SFT-ed models. Additionally, \TPO~enhances the publicly available DeepSeek-R1-Distill-Qwen-7B model, raising its accuracy on the MATH500 dataset from 87.4\% to 91.2\%. These results underscore that \TPO~provides a lightweight solution that improves reasoning capabilities without high resources and \TPO’s ability to overcome performance bottlenecks in multi-epoch SFT with fixed and limited high-quality long-reasoning data. 

\newpage

\section*{Limitations}

\TPO~can further enhance SFT-ed models without requiring additional high-quality long reasoning data. However, since \TPO~is based on the DPO method, it is sensitive to hyperparameters, requiring careful tuning of $\beta$ and learning rate to achieve optimal improvements. 

\bibliography{custom}

\newpage
\appendix
\label{sec:appendix}
\section{Appendix}
\subsection{Evaluating ThinkPO with Different Temperatures}
\label{Evaluating ThinkPO with Different Temperatures}

In our experiments, we initially evaluated the model at a temperature of 0.7. While this provides a good measure of performance, it is important to explore different sampling conditions for a more robust analysis. Therefore, we additionally tested temperatures of 0.1 and 0.5 to examine how ThinkPO impacts Bespoke-Strato-7B under varying levels of randomness in sampling. By comparing results across these temperature settings, we can assess whether ThinkPO consistently enhances the model’s reasoning ability regardless of generation strategy. To provide a comprehensive evaluation, we average the results across all three temperatures. The results are shown in \cref{table: Experiments on different temperatures}.

Our findings demonstrate that ThinkPO consistently improves model performance across different temperature settings. Specifically, at temperatures of 0.1 and 0.7, accuracy increases on four datasets, while at 0.5, improvements are observed on three. To gain a more holistic understanding of ThinkPO’s impact, we average the results across all temperature settings, showing that ThinkPO enhances performance on all five datasets. Notably, on MATH500, ThinkPO improves accuracy by 1.4\%. These results further validate the effectiveness of our proposed method and demonstrate its ability to consistently enhance reasoning performance across different sampling conditions.

\begin{table*}
  \centering
  \caption{
  \label{table: Experiments on different temperatures}
Evaluation of Bespoke-Strato-7B with different temperatures(0.1,0.5,0.7).
Across different values of temperatures, the model achieves accuracy improvements on most datasets. After averaging the results, ThinkPO consistently enhances the model’s performance across all five datasets.
  }\vspace{-10pt}
  \begin{tabular}{c|cc|cc|cc|ccc}
  \toprule
   & \multicolumn{2}{c|}{Temperature=0.1} & \multicolumn{2}{c|}{Temperature=0.5} & \multicolumn{2}{c|}{Temperature=0.7} & \multicolumn{3}{c}{Average} \\
   & {\small+SFT} &  {\small+\textbf{\TPO}} &  {\small+SFT} &  {\small+\textbf{\TPO}} & {\small+SFT} &  {\small+\textbf{\TPO}} & {\small+SFT} &  {\small+\textbf{\TPO}} & Improv. \\
  \midrule
   {MATH500} & 70.2 & 73.4 \upgreen & 81.4 & 82.6\upgreen & 84.0  & 82.8 \downred& 78.5 & 79.6\upgreen & \cellcolor{gray!20}{\color{darkgreen}1.4\%} \\
  {AIME}  &10.0 &  16.7 \upgreen& 20.0 & 16.7\downred  & 20.0 & 23.3\upgreen & 16.7 & 18.9 \upgreen& \cellcolor{gray!20}{\color{darkgreen}13.2\%} \\
    {GPQA} & 34.9 & 30.8\downred & 33.8 & 41.0\upgreen & 37.9 & 43.4\upgreen & 35.5 &  38.4\upgreen& \cellcolor{gray!20}{\color{darkgreen}8.1\%} \\
   {GSM8K} & 89.3 &  91.0 \upgreen & 92.4 &92.3\downred& 92.9 & 93.3 \upgreen& 91.5 & 92.2\upgreen& \cellcolor{gray!20}{\color{darkgreen}0.7\%} \\
   {Olympiad}& 32.8 & 39.6\upgreen & 42.3 &44.8\upgreen & 44.1 & 48.5\upgreen & 39.7& 44.3\upgreen & \cellcolor{gray!20}{\color{darkgreen}11.6\% }\\
   \bottomrule
  \end{tabular}
\end{table*}

\subsection{Analysis of our Reproduce Model in other datasets}
\label{Analysis of our Reproduce Model in other datasets}
\begin{figure*}[t]
\includegraphics[width=0.33\linewidth]{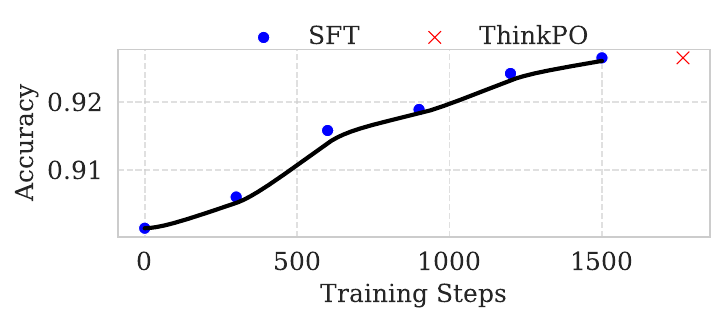}
\includegraphics[width=0.33\linewidth]{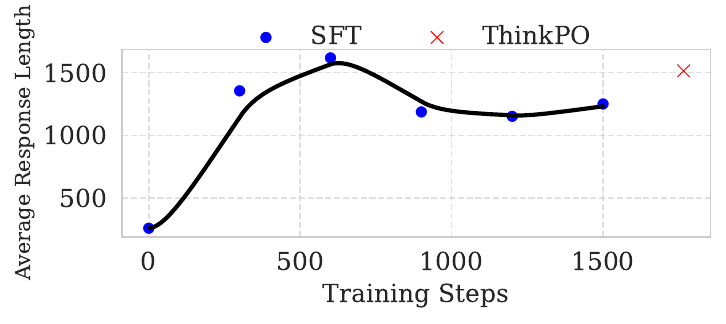}
\includegraphics[width=0.33\linewidth]{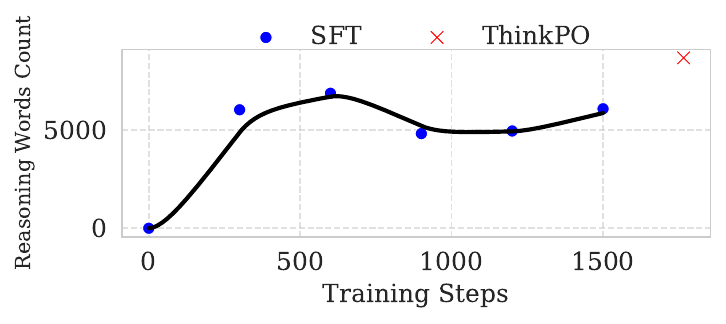}
 \caption{
   Analysis of accuracy(\textbf{Left}), average response length(\textbf{Middle}) and reasoning-supportive words count(\textbf{Right}, like wait, hmm, etc) in reproducing Bespoke-Stratos-7B. We evaluate the model on GSM8K every 300 steps and record results. In the early training stages, all of them improve significantly. However, in the later stages (e.g., after 1200 steps), the model’s performance plateau. When ThinkPO is applied, we see additional improvements in all of the three aspects, demonstrating the effectiveness of Think Preference Optimization.
  }
\hfill

\includegraphics[width=0.33\linewidth]{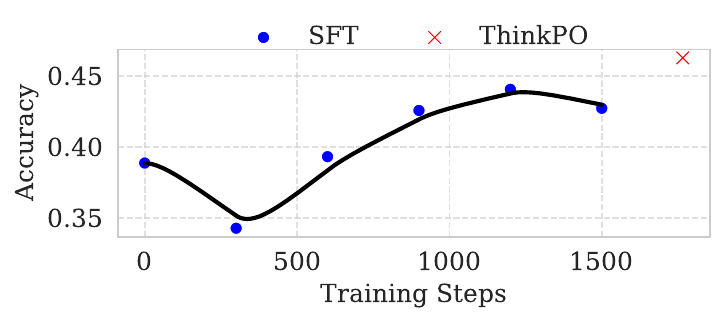}
\includegraphics[width=0.33\linewidth]{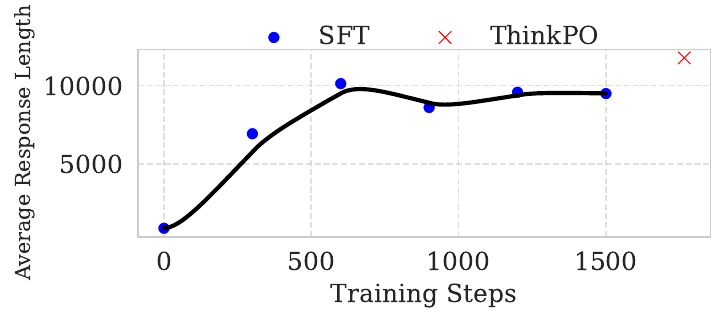}
\includegraphics[width=0.33\linewidth]{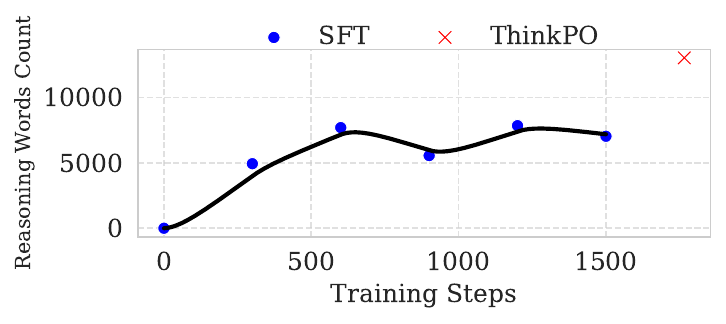}
 \caption{
   Analysis of accuracy(\textbf{Left}), average response length(\textbf{Middle}) and reasoning-supportive words count(\textbf{Right}, like wait, hmm, etc) in reproducing Bespoke-Stratos-7B. We evaluate the model on OlympiadBenchMath every 300 steps and record results. In the early training stages, all of them improve significantly. However, in the later stages (e.g., after 1200 steps), the model’s performance plateau. When ThinkPO is applied, we see additional improvements in all of the three aspects, demonstrating the effectiveness of Think Preference Optimization.
  }\label{fig:sft_length_plot_other}
\end{figure*}

Previously, we only presented the changes in accuracy, average response length, and reasoning-supportive words count over training steps on the MATH500 dataset. Here, we extend our analysis by showcasing results on two additional datasets (like GSM8K) from our reproduced model. The detailed results are illustrated in \cref{fig:sft_length_plot_other}.

As observed in the results for GSM8K and Olympiad Bench Math, the model exhibits a similar trend to MATH500 across all three metrics. During the early stages of SFT, the model’s reasoning ability improves rapidly. However, in later stages, it reaches a performance plateau. ThinkPO effectively helps the model overcome this bottleneck, further enhancing its reasoning capability.

\subsection{Training Recipe}
\label{Training Recipe}

\begin{table*}
  \centering
  \caption{
  The optimal hyperparameters identified in our experiments are listed here, including batch size, learning rate (lr), and beta. These parameters were carefully tuned to achieve the best performance improvements.
  }\label{table:ThinkPO_hyperparameters}\vspace{-10pt}
  \begin{tabular}{c|ccc|ccc|ccc}
  \toprule
   & \multicolumn{3}{c}{Deepseek-7B}& \multicolumn{3}{c}{Bespoke-7B} & \multicolumn{3}{c}{Bespoke-7B-reproduced}  \\
  \midrule
   {batch size}& \multicolumn{3}{c}{48} & \multicolumn{3}{c}{48} & \multicolumn{3}{c}{48}\\
  {lr} & \multicolumn{3}{c}{1e-7} & \multicolumn{3}{c}{5e-7} & \multicolumn{3}{c}{3e-7}  \\
    {$\beta$}& \multicolumn{3}{c}{0.01} & \multicolumn{3}{c}{0.01} & \multicolumn{3}{c}{0.01} \\
   \midrule
    & \multicolumn{3}{c}{Qwen2.5-3B-SFT}& \multicolumn{3}{c}{Qwen2.5-7B-SFT} & \multicolumn{3}{c}{Qwen2.5-14B-SFT}  \\
    \midrule
   {batch size}& \multicolumn{3}{c}{48} & \multicolumn{3}{c}{48} & \multicolumn{3}{c}{48}  \\
  {lr}& \multicolumn{3}{c}{5e-7} & \multicolumn{3}{c}{8e-8} & \multicolumn{3}{c}{1e-7}  \\
    {$\beta$}& \multicolumn{3}{c}{0.01} & \multicolumn{3}{c}{0.01} & \multicolumn{3}{c}{0.01} \\
    \bottomrule
  \end{tabular}
\end{table*}

Here, we provide the corresponding hyperparameters—batch size, learning rate, and $\beta$—that were used to achieve these optimal outcomes. All the hyperparameters are presented in \cref{table:ThinkPO_hyperparameters}.

Besides, we present the training loss curves, gradient norm curves, and margin curves for three models during the \TPO~phase in \cref{fig:dpo trainging loss}. These metrics provide insights into how the models perform throughout the training process, including their convergence behavior, stability of gradients, and the differences in preference between chosen and rejected samples. By examining these curves, we can better understand the effectiveness of \TPO~in enhancing model performance.

\subsection{Examples of LLM's outputs before and after ThinkPO}
\label{Examples of LLM's outputs before and after ThinkPO}
We present the changes in the total number of reasoning-supportive words (such as wait, hmm, let’s think, etc.) throughout both the SFT and ThinkPO training stages in \cref{fig:sft_length_plot} and \cref{fig:sft_length_plot_other}. These words serve as indicators of the model’s reasoning process, reflecting its ability to structure logical steps before arriving at a final answer. Our results show that the number of reasoning-supportive words increases significantly during the initial stages of SFT but eventually plateaus, suggesting that conventional fine-tuning alone may not be sufficient to further enhance structured reasoning. However, after applying ThinkPO, we observe a clear upward trend in the use of these reasoning-supportive expressions, indicating that our method effectively encourages the model to adopt a more deliberative reasoning process.

We provide examples of model outputs before and after applying ThinkPO in \cref{fig:example} and \cref{fig:example2}. Before ThinkPO, the model’s responses tend to be more direct, with fewer reasoning-supportive words, often resulting in incorrect or incomplete answers. In contrast, after applying ThinkPO, the model generates responses that utilize a greater number of reasoning-supportive words. This shift leads to a noticeable improvement in answer correctness, reinforcing the effectiveness of ThinkPO in enhancing the model’s reasoning ability. These findings highlight that ThinkPO not only improves accuracy but also aligns the model’s output with human-like problem-solving patterns.

\begin{table*}[ht]
    \centering
    \resizebox{\linewidth}{!}{
    \small  
    \begin{tabular}{p{0.5in}|p{5.5in}}
        \toprule
        \centering Question & What is the domain of the function $f(x)=\frac{2-x}{log(2-log(x-2))}$, where $log$ is the base $10$ logarithm function? Express your answer in interval notation. \\
        \midrule
        \centering SFT & Okay, so I need to find the domain of the function f(x) = (2 - x)/log(2 - log(x - 2)). \textcolor{red}{hmm}, let's start by recalling that the domain of a function includes all the real numbers x for which the function is defined. Since this function has a logarithm in the denominator, there are a couple of things I need to check: first, the argument of the outer logarithm (the one in the denominator) must be positive because logarithms are only defined for positive numbers. Second, the argument of the inner logarithm (the one inside the denominator's argument) must also be positive because we can't take the logarithm of a non-positive number. Let me break this down step by step.... But \textcolor{red}{\textcolor{red}{wait}}, 102 is already the upper bound, so actually, the domain is (2, 102). But we have to make sure that within (2, 102), the expression 2 - log(x - 2) is positive, which we already considered. Let me verify this again. So, starting with the inner logarithm: x > 2. Then, log(x - 2) is defined. Then, the outer logarithm requires that 2 - log(x - 2) > 0, which as we saw, leads to x < 102. So combining x > 2 and x < 102 gives (2, 102). Additionally, we need to ensure that the denominator is not zero, which occurs at x = 102, but since 102 is already the upper limit, the interval is open at 102. Therefore, the domain is (2, 102). Let me check if there are any other restrictions or if I missed something. \textcolor{red}{wait}, another thing to consider: the numerator is 2 - x. The numerator is a linear function, so it's defined for all real numbers. The denominator is log(2 - log(x - 2)), ... Then, log(3 - 2) = log(1) = 0. Then, 2 - 0 = 2, so log(2) is the denominator. That's valid. If x = 100, then log(100 - 2) = log(98), which is positive, so 2 - log(98) would be 2 - log(98), which is positive because log(98) is less than 2 (since $10^2 = 100$). So, denominator is positive. If x approaches 2 from the right, x approaches 2+, so x - 2 approaches 0+, so log(x - 2) approaches -infty, so 2 - log(x - 2) approaches +infty, so denominator is defined. If x approaches 102 from the left, x approaches 102-, so x - 2 approaches 100-, so log(x - 2) approaches log(100-), which is close to 2, so 2 - log(x - 2) approaches 0 from the positive side, so the denominator approaches 0+, making the whole function approach +infty. But x can't be 102. So, in all other points between 2 and 102, the denominator is defined and positive, so the function is defined there. Therefore, the domain is indeed (2, 102). I think that's it. \textbf{Final Answer} The domain of the function is (2, 102). \\
        \midrule
        \centering ThinkPO & Okay, so I need to find the domain of the function f(x) = (2 - x)/log(2 - log(x - 2)), where the log here is base 10. \textcolor{red}{hmm}, let me think step by step about what's required here.  ... But \textcolor{red}{wait}, I need to check if this x > 2 is compatible with the first condition. So maybe I need to solve both inequalities together?Let me start with the second condition: log(x - 2) is defined when x - 2 > 0, so x > 2. So that's the domain from the argument of the inner logarithm. But then, ... So combining x > 2 and x < 102 gives me 2 < x < 102. \textcolor{red}{wait}, but hold on. Let me verify this. If log(x - 2) < 2, then $x - 2 < 10^2 = 100$, so x < 102. But since x - 2 must be positive (because log requires positive argument), x > 2. So indeed, x is in (2, 102). But \textcolor{red}{wait}, when I take log(x - 2) < 2, ... If x approaches 2 from the right, x - 2 approaches 0+, so log(x - 2) approaches -infty, so 2 - log(x - 2) approaches infinity. \textcolor{red}{wait}, but as x approaches 2+, the inner log(x - 2) goes to -infty, so 2 - log(x - 2) goes to +infty. Therefore, the denominator log(2 - log(x - 2)) approaches log(infinity) which is +infty. \textcolor{red}{wait}, but that would mean the denominator is positive and approaching infinity. .... But \textcolor{red}{wait}, is there a point where the denominator could be zero? Let's check. For 2 - log(x - 2) = 0, then log(x - 2) = 2, so $x - 2 = 10^2 = 100$, so x = 102. But x = 102 is the upper bound of the domain. So at x = 102, the denominator is log(2 - log(100)) = log(2 - 2) = log(0), which is undefined. So x cannot be 102. So the domain is (2, 102). But \textcolor{red}{wait}, we also have to consider the numerator. The numerator is 2 - x. So 2 - x can be zero, but if the denominator is non-zero, then the function is defined. If numerator is zero, then f(x) = 0/denominator, which is zero. But division by zero is undefined. \textcolor{red}{wait}, but if numerator is zero, does that make the entire expression zero, but with denominator non-zero? For example, if x = 2, then numerator is 0, denominator is log(2 - log(0)). \textcolor{red}{wait}, x=2 is not in the domain because x - 2 = 0, so log(0) is undefined. ...\textcolor{red}{wait}, but the denominator is log(2 - log(x - 2)). So the denominator is log of something. The logarithm output can be any real number, but in this case, the denominator is used in the function, so as long as the argument inside the log is positive, which we already considered. But the denominator itself can be any real number except zero. \textcolor{red}{wait}, no. \textcolor{red}{wait}, the denominator is log(2 - log(x - 2)), ....\textcolor{red}{wait}, so...Let's compute log(10) is 1 (base 10), so 2 - 1 = 1, log(1) = 0, ...That seems correct.\textcolor{red}{wait}, but let's check another value. If x is 11, then $x - 2 = 9, log(9) = 0.954, 2 - 0.954 = 1.046$, so log(1.046) is a positive number. So denominator is defined and non-zero. If x is 13, then x - 2 = 11, log(11) = 1.041, 2 - 1.041 = 0.959, log(0.959) is negative. But \textcolor{red}{wait}, the original function is defined only if the denominator is log(positive number). So if the denominator is log of a number less than 1, that's negative, but is that allowed? \textcolor{red}{wait}, no. \textcolor{red}{wait}, the denominator is log(2 - log(x - 2)). For the log to be defined, 2 - log(x - 2) must be positive. So if 2 - log(x - 2) is negative, then log of that is undefined (since log of a negative number is not real). .... \textcolor{red}{wait}, x=1 is less than 2,... \textcolor{red}{wait}, no. \textcolor{red}{wait}, the denominator is log(2 - log(x -2)). For log to be defined, the argument must be positive. So if 2 - log(x -2) is positive, then log is defined, regardless of whether it's positive or negative. \textcolor{red}{wait}, no. \textcolor{red}{wait}, .... So intervals (2,12) and (12,102). That should be correct. \\
        \bottomrule
    \end{tabular}
    }
     \caption{One example of LLM's wrong outputs before ThinkPO and correct outputs after ThinkPO. ThinkPO helps models generate more reasoning-supportive discourse markers, such as wait, hmm, and other hesitation cues. }
    \label{fig:example}
\end{table*}

\begin{table*}[ht]
    \centering
    \resizebox{\linewidth}{!}{
    \small  
    \begin{tabular}{p{0.5in}|p{6.4in}}
        \toprule
        \centering Question & How many positive two-digit integers are factors of both 100 and 150? \\
        \midrule
        \centering SFT & Okay, so I need to figure out how many positive two-digit integers are factors of both 100 and 150. \textcolor{red}{hmm}, let's start by understanding the problem....So, the GCD of 100 and 150 is 50. That means all the common factors of 100 and 150 are the factors of 50. Let me list out the factors of 50 to make sure. The factors of 50 are 1, 2, 5, 10, 25, and 50. \textcolor{red}{wait}, that's six factors. But the problem asks for two-digit integers. From this list, the two-digit factors would be 25 and 50. So that's two numbers. But \textcolor{red}{wait}, hold on. Let me verify if 25 and 50 are indeed factors of both 100 and 150. For 25: 100 / 25 = 4, which is an integer. 150 / 25 = 6, which is also an integer. So yes, 25 is a common factor. For 50: 100 / 50 = 2, and 150 / 50 = 3. Both integers, so 50 is also a common factor. But \textcolor{red}{wait}, are there any other two-digit numbers that are factors of both 100 and 150? Let's check. The factors of 100 are 1, 2, 4, 5, 10, 20, 25, 50, 100. The factors of 150 are 1, 2, 3, 5, 6, 10, 15, 25, 30, 50, 75, 150. The common factors are 1, 2, 5, 10, 25, 50. So, the two-digit common factors are 25 and 50. That gives us two numbers. \textcolor{red}{wait}, but hold on. The problem says "positive two-digit integers". ... So, factors of 50 are 1, 2, 5, 10, 25, 50. So, two-digit ones are 25 and 50. Therefore, the answer is 2. But \textcolor{red}{wait}, let me check if there are any other numbers between 10 and 99 that divide both 100 and 150. For example, 10 divides both. 10 is two-digit. 20 divides 100 but 150 / 20 is 7.5, which is not integer. So 20 is out. 30 divides 150 but 100 / 30 is approximately 3.333, not integer. 15 divides 150 but 100 / 15 is about 6.666, not integer. 75 divides 150 but 100 / 75 is 1.333, not integer. So, indeed, only 25 and 50 are in the two-digit range. Therefore, the answer is 2. But \textcolor{red}{wait}, \textcolor{red}{wait} a second. Let me check 25 and 50 again. 25 divides both 100 and 150, yes. 50 divides both, yes. Are there any other numbers between 10 and 99 that divide both? Let me check 10. 10 divides 100 and 150. 10 is two-digit. But we already have 10, 25, 50. \textcolor{red}{wait}, 10 is a factor, but it's single-digit. So 10 is not included. So 25 and 50 are the only two. Therefore, the answer is 2. So, the number of positive two-digit integers that are factors of both 100 and 150 is 2. But just to be thorough, let me check 25: 100 / 25 = 4, 150 / 25 = 6. Both integers. 50: 100 / 50 = 2, 150 / 50 = 3. Both integers. So yes, they are factors. Are there any others? Let me check 30: 100 / 30 = 3.333, not integer. 40: 100 / 40 = 2.5, not integer. 45: 100 / 45 = 2.222, no. 55: 100 / 55 = 1.818, no. 60: 100 / 60 = 1.666, no. 65: 100 / 65 = 1.538, no. 70: 100 / 70 = 1.428, no. 75: 100 / 75 = 1.333, no. 80: 100 / 80 = 1.25, no. 85: 100 / 85 = 1.176, no. 90: 100 / 90 = 1.111, no. 95: 100 / 95 = 1.052, no. So, none of these two-digit numbers are factors of both 100 and 150. Therefore, only 25 and 50. Therefore, the answer is 2. So, the final answer is 2, which in boxed form is 2. \\
        \midrule
        \centering ThinkPO & Okay, so I need to figure out how many positive two-digit integers are factors of both 100 and 150. \textcolor{red}{hmm}, let's start by understanding the problem.... For prime 5, both have exponent 2, so min(2,2)=2. Prime 3 is only in 150, so we ignore it. So GCD is 2 * 5 * 5 = 2 * 25 = 50. \textcolor{red}{wait}, so the GCD of 100 and 150 is 50. ... \textcolor{red}{wait}, that's two numbers. But let me verify. \textcolor{red}{wait}, 50's divisors: 1, 2, 5, 10, 25, 50. Yep, that's all. So two-digit numbers here are 10 and 50. So the answer would be 2? But let me make sure I didn't miss anything. But \textcolor{red}{wait}, is 10 and 50 the only two-digit divisors of 50? Let's check. 50 divided by 10 is 5, which is integer. 50 divided by 50 is 1. So yes, those are the divisors. But \textcolor{red}{wait}, 5 is a single-digit number. So the two-digit divisors are 10 and 50. So that's two numbers. \textcolor{red}{wait}, but the problem says \"positive two-digit integers\", so 10 and 50. But \textcolor{red}{wait}, is 50 a two-digit number? Yes, 50 is two digits. But 10 is also two digits. So that's two numbers. So the answer is 2? \textcolor{red}{hmm}, but let me cross-verify this. Alternatively, maybe I should approach it another way. ...So from 10 to 99, the numbers that divide both 100 and 150 are 10, 25, and 50. \textcolor{red}{wait}, but 25 and 50 are two-digit numbers. So that's three numbers: 10, 25, 50. \textcolor{red}{wait}, but why did I think earlier that 25 is a two-digit number? It is. So three numbers. But then 10, 25, 50. So three numbers. But when I first thought of the GCD approach, I considered the divisors of 50, which are 1, 2, 5, 10, 25, 50. So only two two-digit divisors. But here, when checking manually, I found three. So there's a discrepancy. Which one is correct?\textcolor{red}{wait}, perhaps I made a mistake in the initial approach. Let's check again. The GCD of 100 and 150 is 50. So the common factors are the divisors of 50. Divisors of 50 are 1, 2, 5, 10, 25, 50. So yes, two-digit divisors are 10 and 50. So that's two numbers. But according to the manual check, there are three. So which is correct?\textcolor{red}{wait}, let's check 25. 25 is a divisor of both 100 and 150. 25 divides 100 because 25*4=100, and 25 divides 150 because 25*6=150. So 25 is indeed a common factor. But 25 is a two-digit number, right? 2 and 5. So yes, two digits. So 25 is a two-digit divisor. Then 10 and 50. So that's three numbers. But according to the GCD method, the common factors are exactly the divisors of 50, which are six numbers, but only two two-digit. However, when I checked manually, I found three two-digit divisors. So where's the mistake?\textcolor{red}{wait}, maybe I made a mistake in the GCD calculation. Let me check again. 100 factors: $2^2 * 5^2$. 150 factors: $2^1 * 3^1 * 5^2$. Common prime factors: $2^min(2,1)=1$, $5^min(2,2)=2$. So GCD is $2^1 *5^2=2*25=50$. So that's correct. So the common factors are exactly the divisors of 50. Therefore, divisors of 50 are 1,2,5,10,25,50. So the two-digit ones are 10,25,50. \textcolor{red}{wait}, that's three numbers. But according to GCD approach, divisors of GCD. So why the discrepancy? \textcolor{red}{wait}, but if the GCD is 50, then the common factors are exactly the divisors of 50. So that's six numbers, but only three are two-digit. So why did the manual check show three? But according to the list, divisors of 50 are 1,2,5,10,25,50. So two-digit are 10,25,50. That's three. So that's correct. But why did I think earlier that there were two? Because maybe I made a mistake in the first thought process. \textcolor{red}{wait}, no. \textcolor{red}{wait}, when I first considered the problem, I thought the answer is two. But then when I did the manual check, I found three. So which is correct? Let me verify each of the three numbers:10: 10 divides 100 (10*10=100) and 150 (10*15=150). So yes.25: 25*4=100, 25*6=150. Yes, so 25 divides both.50: 50*2=100, 50*3=150. Yes, so 50 divides both.So all three are valid. Therefore, the answer should be 3. But \textcolor{red}{wait}, the initial GCD approach said divisors of 50, which are six numbers, but only three are two-digit. However, when I checked the manual approach, I found three. Therefore, the answer is three? But I need to make sure there are no other two-digit numbers that divide both 100 and 150. \textcolor{red}{wait}, let's check 30. 100 divided by 30 is 3.333, not integer. .... But \textcolor{red}{wait}, let me check another way. Alternatively, since the GCD is 50, the common factors are the divisors of 50. So divisors of 50 are 1,2,5,10,25,50. So two-digit are 10,25,50. So three numbers. Therefore, the answer is 3. But why did I initially think it was two? Maybe I confused divisors of 50 with common divisors. But no, the common divisors are exactly the divisors of the GCD. So if GCD is 50, then common divisors are divisors of 50. Therefore, three numbers. \textcolor{red}{wait}, but in the initial thought process, I considered that the GCD is 50, so common factors are 1,2,5,10,25,50. Then two-digit are 10,25,50. Three numbers. Therefore, the answer is 3. But let me check once more. Let's take 25. 25 divides 100 and 150. 25*4=100, 25*6=150. Yes. 10: 10*10=100, 10*15=150. Yes. 50: 50*2=100, 50*3=150. Yes. So all three are valid. Therefore, the number of positive two-digit integers that are factors of both 100 and 150 is 3. But \textcolor{red}{wait}, hold on. Let me check 5. 5 is a divisor of both, but it's single-digit. 1, 2, 5 are also divisors but single-digit. So the only two-digit divisors are 10,25,50. So three numbers. Therefore, the final answer should be 3. But \textcolor{red}{wait}, when I first did the GCD approach, I thought it was two, but after manual checking, I found three. So why the discrepancy? ..... But \textcolor{red}{wait}, the initial problem says \"positive two-digit integers\", so numbers from 10 to 99. So 10,25,50 are within that range. So yes, three numbers. Therefore, after thorough checking, the answer is 3. So I need to make sure I didn't make a mistake in the initial GCD approach. But since GCD is 50, and common factors are divisors of 50, then three two-digit. So the answer is 3. But let me also check if 50 is indeed a two-digit number. Yes, 50 has two digits. 10 is two-digit. 25 is two-digit. So three numbers. Therefore, I think the correct answer is 3. So the final answer should be 3, boxed as 3\\
        \bottomrule
    \end{tabular}
    }
     \caption{One more examples of LLM's wrong outputs before ThinkPO and correct outputs after ThinkPO. ThinkPO helps models generate more reasoning-supportive discourse markers, such as wait, hmm, and other hesitation cues. }
    \label{fig:example2}
\end{table*}

\end{document}